\documentclass[preprint,5p,times,authoryear,twocolumn]{elsarticle}

\usepackage{amssymb}
\usepackage{amsmath}
\usepackage{hyperref}
\usepackage{url}
\usepackage{array}
\usepackage{booktabs}
\usepackage{threeparttable}
\usepackage{caption}
\usepackage{subcaption}
\usepackage{longtable}
\usepackage{lineno}

\usepackage{makecell}
\usepackage{multirow}
\usepackage{tikz}
\usetikzlibrary{shapes,arrows,positioning}
\usepackage{enumitem}

\definecolor{darkgreen}{rgb}{0.0, 0.5, 0.0}
\newcommand{\impr}[1]{\scalebox{0.6}{\scriptsize\textcolor{darkgreen}{(+#1\%)}}}

\newcommand{\nimpr}[1]{\scalebox{0.6}{\scriptsize\textcolor{red}{(-#1\%)}}}

\hypersetup{
    colorlinks=true,
    urlcolor=magenta,      
    linkcolor=magenta,     
    citecolor=magenta,     
}

\journal{}

\begin{document}

\begin{frontmatter}

\title{Scalable Graph Attention‐Based Instance Selection via Mini‐Batch Sampling and Hierarchical Hashing}

\author[uaeu,ku]{Zahiriddin Rustamov}  
\author[uaeu_math]{Ayham Zaitouny}
\author[uaeu]{Nazar Zaki\corref{cor1}}  

\cortext[cor1]{Corresponding author. Email: nzaki@uaeu.ac.ae, Phone: +971-37672018}

\affiliation[uaeu]{organization={College of Information Technology, United Arab Emirates University},
            city={Al Ain},
            postcode={15551}, 
            country={United Arab Emirates}}

\affiliation[ku]{organization={Department of Computer Science, KU Leuven},
            country={Belgium}}

\affiliation[uaeu_math]{organization={College of Science, United Arab Emirates University},
            city={Al Ain},
            postcode={15551}, 
            country={United Arab Emirates}}

\begin{abstract}

Instance selection (IS) addresses the critical challenge of reducing dataset size while keeping informative characteristics, becoming increasingly important as datasets grow to millions of instances. Current IS methods often struggle with capturing complex relationships in high-dimensional spaces and scale with large datasets. This paper introduces a graph attention-based instance selection (GAIS) method that uses attention mechanisms to identify informative instances through their structural relationships in graph representations. We present two approaches for scalable graph construction: a distance-based mini-batch sampling technique that achieves dataset-size-independent complexity through strategic batch processing, and a hierarchical hashing approach that enables efficient similarity computation through random projections. The mini-batch approach keeps class distributions through stratified sampling, while the hierarchical hashing method captures relationships at multiple granularities through single-level, multi-level, and multi-view variants. Experiments across 39 datasets show that GAIS achieves reduction rates above 96\% while maintaining or improving model performance relative to state-of-the-art IS methods. The findings show that the distance-based mini-batch approach offers an optimal efficiency for large-scale datasets, while multi-view variants excel on complex, high-dimensional data, demonstrating that attention-based importance scoring can effectively identify instances important for maintaining decision boundaries while avoiding computationally prohibitive pairwise comparisons.

\end{abstract}

\begin{keyword}
instance selection \sep graph attention networks \sep mini-batch sampling \sep locality-sensitive hashing \sep graph data selection
\end{keyword}

\end{frontmatter}



\section{Introduction}\label{sec:introduction}

Modern machine learning (ML) pipelines can lead to high computational costs with limited performance gains when trained on large datasets that contain redundant or less informative instances. In datasets with millions of examples, training times may extend to several minutes or hours, while the model performance often levels off well before using all available data. Instance selection (IS) tackles this issue by identifying and preserving only the most informative instances while removing redundant data points, which cuts down computational needs while keeping model performance within acceptable levels \citep{Cunha2023}.

Formally, given a dataset \( \mathcal{D} = \{(x_i, y_i)\}_{i=1}^{n} \), where \( x_i \in \mathbb{R}^d \) and \( y_i \in \{1, \ldots, C\} \), IS aims to select a subset \( \mathcal{S} \subseteq \mathcal{D} \) that reduces the number of instances while ensuring the classifier's performance stays within an acceptable range compared to training on the full dataset:
\[
\mathcal{S}^* = \arg \min_{\mathcal{S} \subseteq \mathcal{D}} |\mathcal{S}| \quad \text{subject to} \quad f(\mathcal{S}) \geq \alpha\ \cdot f(\mathcal{D})
\]
where \( f(\mathcal{S}) \) is the performance metric of a model \( h_{\mathcal{S}} \) trained on \( \mathcal{S} \) and evaluated on a test set \( \mathcal{T} \), and \( \alpha \in (0, 1) \) specifies the minimum acceptable relative performance.

Traditional IS methods rely mostly on nearest neighbor computations, falling into three groups: condensation (keeping instances near decision boundaries), edition (removing noisy instances), and hybrid approaches \citep{Cano2003, Garcia2012, Saha2022}. While these work well for moderately sized datasets, they struggle to capture complex relationships in high-dimensional spaces. Their need for pairwise distance calculations results in quadratic complexity that becomes prohibitive for large-scale datasets, often taking several hours of preprocessing time on datasets with millions of instances.

Graph-based methods offer a promising alternative by representing data relationships through node connections \citep{Goyal2018}. These approaches can effectively capture structural information through proximity graphs and neighborhood networks, revealing dataset patterns that distance-based methods might miss. Recent advances in graph sampling and reduction \citep{Chen2022} show their potential to handle large-scale computational challenges while retaining key data characteristics.

However, current graph-based methods have limits for IS tasks. Graph reduction techniques change the original graph structure through node merging or edge alterations, losing the direct link to original instances. While graph sampling techniques keep instance identity, the cost of constructing graphs from large non-graph datasets (typically \(O(n^2)\) complexity) remains a significant bottleneck, especially as datasets grow beyond millions of instances.

Graph Attention Networks (GATs) \citep{Velickovic2017} present an unexplored direction for IS through their ability to learn importance weights for node neighborhoods dynamically. The attention mechanisms in GATs can compute instance importance based on local structural properties, providing an interpretable approach to identify informative data points. However, two questions remain: (\textbf{1}) How can we efficiently construct meaningful graph representations from large tabular datasets? (\textbf{2}) How can we utilize attention mechanisms to identify truly informative instances while keeping computational costs feasible?

We address these questions through two approaches: (\textbf{1}) a distance-based mini-batch sampling technique that handles large datasets while preserving class distributions, and (\textbf{2}) a family of locality-sensitive hashing (LSH) methods that enable scalable graph construction through random projections, capturing relationships at multiple granularities.

The key contributions of this work are:
\begin{itemize}[noitemsep,nolistsep]
    \item Development of GAIS (Graph Attention-based Instance Selection), a novel framework that uses attention mechanisms to identify informative instances in tabular data.
    \item Introduction of mini-batch sampling approach that reduces graph construction complexity from $O(n^2)$ to $O(K \times w_{\max}^2)$ while preserving class distributions through strategic sampling with computational bounds.
    \item Design of multi-level and multi-view LSH variants that capture relationships at different granularities through random projections.
    \item Extensive empirical evaluation demonstrating significant improvements over state-of-the-art IS methods across diverse datasets, with particular effectiveness in complex, high-dimensional problems.
\end{itemize}

The rest of the paper is organized as follows: Section \ref{sec:literature_review} reviews related work. Section \ref{sec:methodology} details our proposed methods. Section \ref{sec:experimental_setup} describes the experimental setup. Section \ref{sec:results} presents and discusses our findings. Section \ref{sec:conclusions} concludes with future directions.

\section{Related work}\label{sec:literature_review}

\subsection{Instance Selection}\label{subsec:lr_IS}
Traditional IS methods primarily rely on nearest neighbor computations to evaluate instance relationships and importance, falling into three main categories: condensation (retaining instances near decision boundaries), edition (removing noisy or borderline instances), and hybrid approaches \citep{Garcia2012}. While various improvements have been proposed, including density-based and scalability-focused approaches \citep{Carbonera2018, Garcia-Osorio2010}, recent work has introduced several innovative methodologies that enhance or reformulate how instance relationships are evaluated.

Density-based approaches have evolved from local to global analysis, as demonstrated by GDIS and EGDIS \citep{Malhat2020}, which evaluate instance importance through global density patterns and neighbor relationships. The ranking-based RIS algorithm \citep{Cavalcanti2020} introduces a scoring mechanism that considers relationships between instances, using normalized exponential transform of distances weighted by class membership.
More theoretically grounded approaches such as BDIS \citep{Chen2022b} employ Bayesian decision theory to categorize instances as reducible, irreducible, or deleterious, while leveraging percolation theory to identify local homogeneous clusters---groups of instances with the same label connected by nearest neighbors, where the cluster size reflects the probability of class consistency. Alternative frameworks like CIS \citep{Moran2022} employ reinforcement learning and clustering, where an agent learns to select valuable instance clusters through Q-learning and curiosity-driven exploration.
Recent work has explored novel theoretical frameworks, including approval-based multi-winner voting \citep{SanchezFernandez2024}, which models instances as both voters and candidates in an election system, and evidence theory-based approaches like EIS \citep{Gong2021}, which evaluates instance importance through mass functions and Dempster's rule of combination.

Despite these advances, the fundamental challenge of nearest neighbor computations remains, with typical $O(n^2)$ computational complexity becoming prohibitive for very large datasets, and performance often deteriorating in high-dimensional spaces \citep{Qiao2018, Boutet2016}.
These challenges have motivated the development of various alternative approaches. Among these, graph-based methods have emerged as one promising direction for capturing complex instance relationships, though they present their own unique computational considerations.


\subsection{Graph-based Selection}\label{subsec:lr_graph_IS}
Graph approaches leverage structural information of data by representing data points as nodes and their relationships as edges. In the context of IS, these approaches can be broadly categorized into two main strategies: graph sampling, which selects nodes directly from the original graph, and graph reduction, which modifies the graph structure to create a smaller representation.

\begin{figure}[!t]
    \centering
    \includegraphics[width=0.8\linewidth]{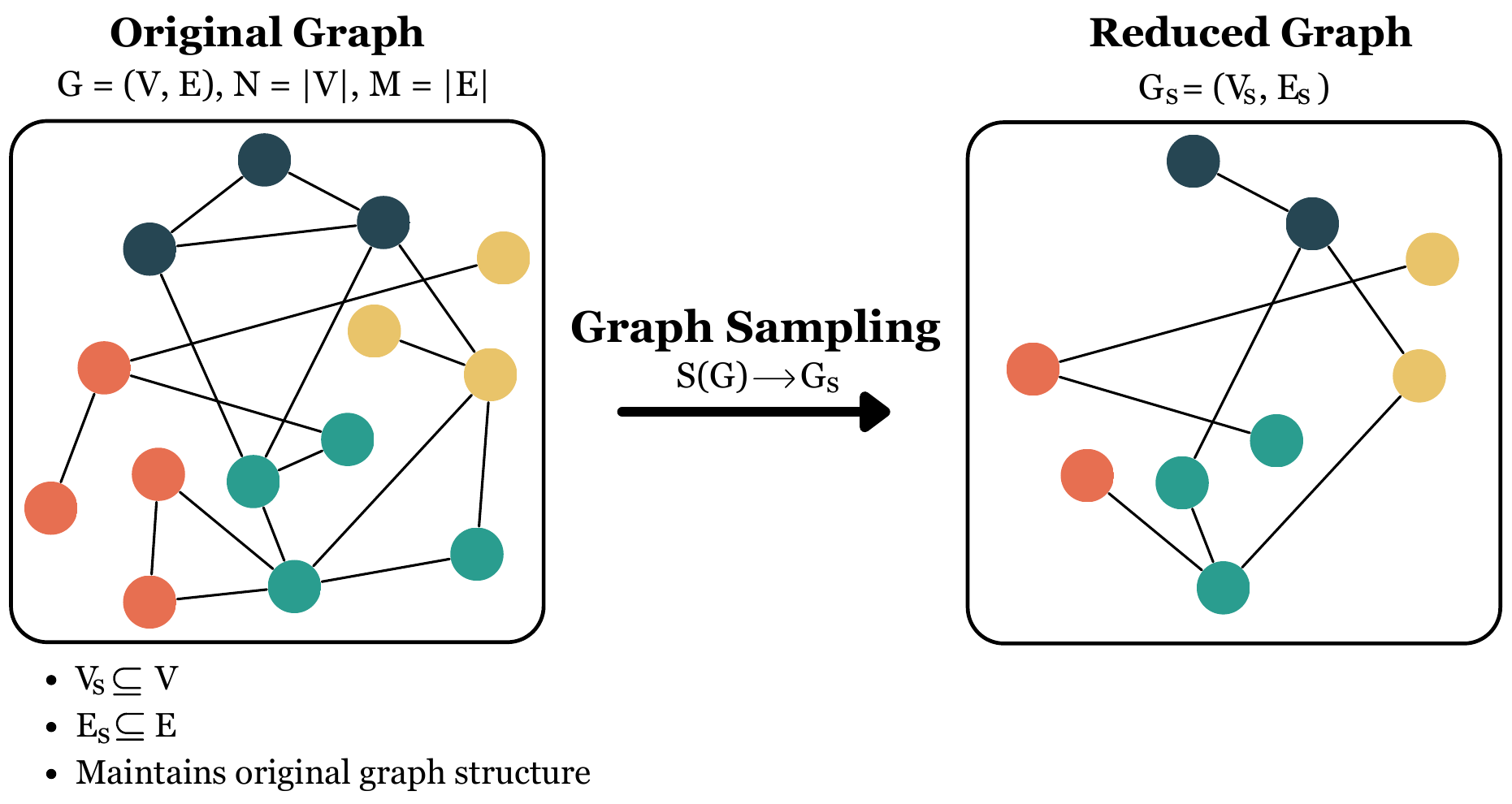}
    \caption{Illustration of a graph sampling process. The original graph $G = (V, E)$ with $|V| = N$ and $|E| = M$ is sampled to produce $G_s = (V_s, E_s)$ where $V_s \subseteq V$, $E_s \subseteq E$, $|V_s| < N$, and $|E_s| < M$. This method selectively retains nodes and edges to create a subgraph approximating the original structure and properties.}
    \label{fig:graph_sampling}
\end{figure}

Graph sampling methods select a subset of nodes from the original graph while preserving its structure \citep{Hu2013}. As illustrated in Figure \ref{fig:graph_sampling}, this process creates a subgraph that maintains the original relationships between data points, making it particularly suitable for IS since selected nodes directly correspond to original instances. Key developments in graph-based IS include proximity graphs for prototype selection \citep{Sanchez1997}, where edges connect instances within a specified distance threshold, Hit Miss Networks that encode nearest neighbor relations \citep{Marchiori2008}, and natural neighborhood graphs for identifying and removing noisy or redundant points \citep{Yang2018}.

\begin{figure}[!t]
    \centering
    \includegraphics[width=0.8\linewidth]{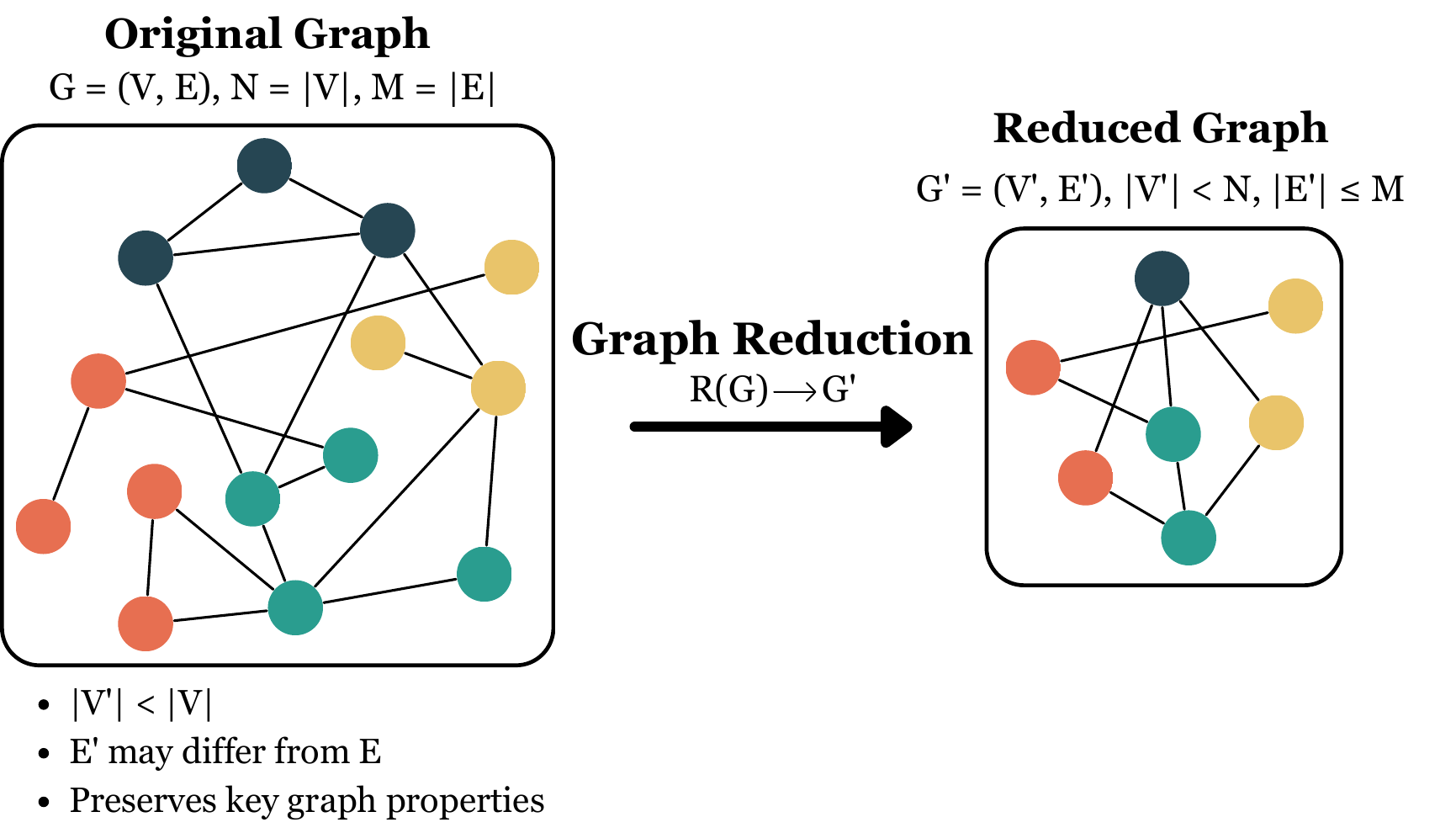}
    \caption{Illustration of a graph reduction process. The original graph $G = (V, E)$ with $|V| = N$ and $|E| = M$ undergoes a reduction to form the reduced graph $G' = (V', E')$ where $|V'| < N$. This process involves condensing or merging nodes and potentially modifying edges to reduce complexity while preserving key graph properties.}
    \label{fig:graph_reduction}
\end{figure}

Graph reduction techniques \citep{Chen2022}, while successful in general graph summarization tasks, have seen limited application in IS. As shown in Figure \ref{fig:graph_reduction}, these methods merge nodes or modify edge structures. When applied to IS, these approaches typically require mapping reduced nodes back to original instances, often by selecting representatives from merged clusters. This indirect correspondence makes these methods less suitable for IS tasks compared to graph sampling approaches that maintain direct relationships to the original instances.

The effectiveness of graph-based methods for IS has been demonstrated by \citet{Rustamov2024}, who evaluated multiple graph-based approaches across diverse classification datasets, showing their potential to achieve substantial data reduction while maintaining model performance.
In our previous work \citep{Rustamov2024GAIS}, we explored using GATs for IS. While this approach showed promise in identifying important instances through attention mechanisms, it was limited by its reliance on confidence scores as importance measures and faced significant scalability issues with large datasets due to its inefficient chunking-based graph construction that still required substantial memory overhead. In this paper, we address these limitations by introducing efficient graph construction methods—distance-based mini-batching and LSH approaches—that significantly reduce computational complexity, alongside a dual-perspective importance scoring mechanism that leverages attention weights to more effectively identify informative instances.

These challenges highlight broader limitations in current large-scale graph-based IS methods. The computational cost and memory requirements for graph construction and training become prohibitive with growing dataset sizes, especially when using distance-based metrics. Current approaches struggle to efficiently analyze entire graph structures in large datasets, particularly when dealing with complex patterns like multi-scale dependencies and non-linear interactions. While graph-based approaches have demonstrated success in various data selection tasks --- from representative subset selection \citep{mall2014} to data pruning \citep{maharana2024} --- their application to IS for tabular data remains underdeveloped, particularly in efficiently handling large-scale datasets with heterogeneous feature relationships and varying data distributions.


\subsection{Scaling Challenges in Graph-based Methods}\label{subsec:lr_graph_scaling}

While graph-based methods show promise for IS, they face significant scaling challenges as datasets grow in size and dimensionality. The primary bottleneck lies in the graph construction phase, particularly for k-nearest neighbor (KNN) graphs that form the foundation of many graph-based approaches. Constructing KNN graphs requires computing pairwise distances between instances, resulting in $O(n^2)$ computational complexity and substantial memory requirements for storing the resulting adjacency matrices \citep{Li2024}.

To address these construction challenges, approximate methods such as Locality-Sensitive Hashing (LSH) have been explored, offering sublinear time complexity \citep{Jafari2021}. LSH works by projecting data points into lower-dimensional hash codes where similar points are likely to share the same hash bucket, enabling faster similarity comparisons. Notable advancements include VRLSH \citep{Eiras-Franco2020}, which uses voted random projections, and the fast LSH-based KNN graph construction algorithm by \citet{Zhang2013}, which significantly improves speed while maintaining accuracy.

However, LSH-based approaches face their own challenges, particularly with high-dimensional tabular data. The fundamental issue is that as dimensionality increases, the probability of hash collisions (different data points sharing the same hash code) becomes less correlated with actual data similarity. This diminished correlation leads to less accurate graph structures \citep{Gu2013}, where edges may connect dissimilar points while missing truly similar ones. Traditional LSH methods also struggle with non-uniform data distributions, where dense regions experience excessive collisions while sparse regions lack sufficient connections, resulting in suboptimal graph constructions that can adversely affect IS.

Recent developments, such as LayerLSH \citep{Ding2022}, address skewed data distributions using a multi-layered index structure, but their effectiveness for graph construction in IS tasks remains unexplored. These challenges suggest that multi-level or hierarchical LSH approaches could offer potential solutions by capturing data relationships at multiple resolutions, potentially addressing both fine-grained local structures and broader global patterns.

\begin{figure*}[!t]
    \centering
    \includegraphics[width=0.9\textwidth]{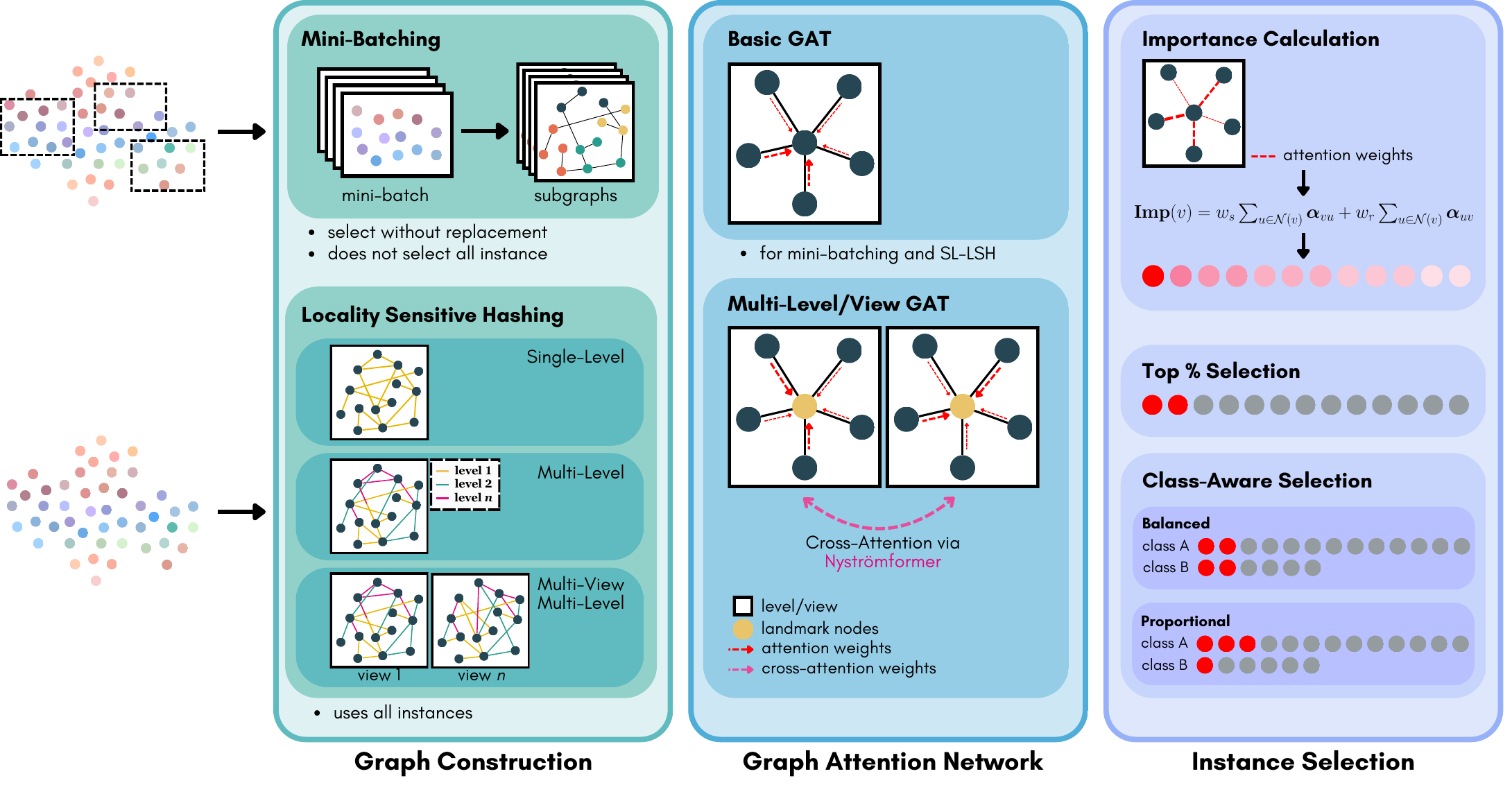}
    \caption{High-level overview of the Graph Attention-based Instance Selection (GAIS) methodology. The framework operates in three stages: (1) \textbf{Graph Construction} - creates scalable graph representations using either distance-based mini-batch sampling that selects instances without replacement to form subgraphs, or Locality-Sensitive Hashing with single-level, multi-level, and multi-view variants that use all instances; (2) \textbf{Graph Attention Network} - applies basic GAT for mini-batching/single-level LSH and enhanced multi-level/multi-view GAT with cross-attention via Nystr\"om approximation to learn node importance through attention weights; (3) \textbf{Instance Selection} - computes importance scores from attention weights and applies either top-percentage global selection or class-aware selection (balanced or proportional) to retain the most informative instances.}
    \label{fig:methodology}
\end{figure*}

Beyond graph construction, computational challenges persist in the subsequent model training phase. Training GNNs on dense graphs requires substantial computational resources for message passing between nodes. Also, dense graphs can result the over-smoothing problem in GNNs \citep{Rusch2023}, where repeated message passing causes node representations to converge to similar values. This over-smoothing effect leads to a loss of discriminative power after multiple layers, potentially undermining the GNN's ability to identify truly informative instances for selection.


\subsection{Gap}\label{subsec:gap}

Despite significant advances in IS methods, a critical gap persists in scalable, efficient graph-based approaches.
First, existing approaches struggle with the computational complexity of graph construction, which typically scales quadratically with dataset size. Even approximate KNN-based methods become prohibitively expensive for datasets exceeding millions of instances, limiting their practical application.
Second, while LSH techniques offer potential scalability improvements, they face challenges in capturing meaningful relationships in high-dimensional spaces, where hash collisions become less correlated with actual data similarity. Traditional LSH implementations often fail to adapt to non-uniform data distributions and struggle to simultaneously represent both fine-grained local structures and broader global patterns.
Third, current graph-based selection methods lack adaptive mechanisms to identify and prioritize truly informative instances. Most approaches either treat all instances with equal importance or rely on static metrics that cannot capture the varying contributions of instances to decision boundaries across different regions of the feature space.

To address these limitations, we propose GAIS which leverages attention mechanisms to identify informative instances through adaptive graph-based representation learning. Our distance-based mini-batch method employs strategic sampling to efficiently process large datasets while preserving class distributions. Our family of LSH methods, including single-level, multi-level, and multi-view variants, enables scalable graph construction that captures relationships at multiple granularities. By utilizing attention weights as importance indicators, GAIS adaptively prioritizes instances that contribute most to decision boundaries, even with approximated graph structures.


\section{Methods}\label{sec:methodology}

This section outlines GAIS's methodology, covering graph construction, GAT implementation, and IS. Figure~\ref{fig:methodology} illustrates the overall process.

\subsection{Graph Construction}
We propose two graph construction methods: distance-based with mini-batch sampling and hashing-based using LSH with multi-level and multi-view structures.

\subsubsection{Distance-based Mini-Batching Method}\label{subsubsec:distance_based_method}

\begin{figure*}[!t]
    \centering
    \includegraphics[width=0.8\textwidth]{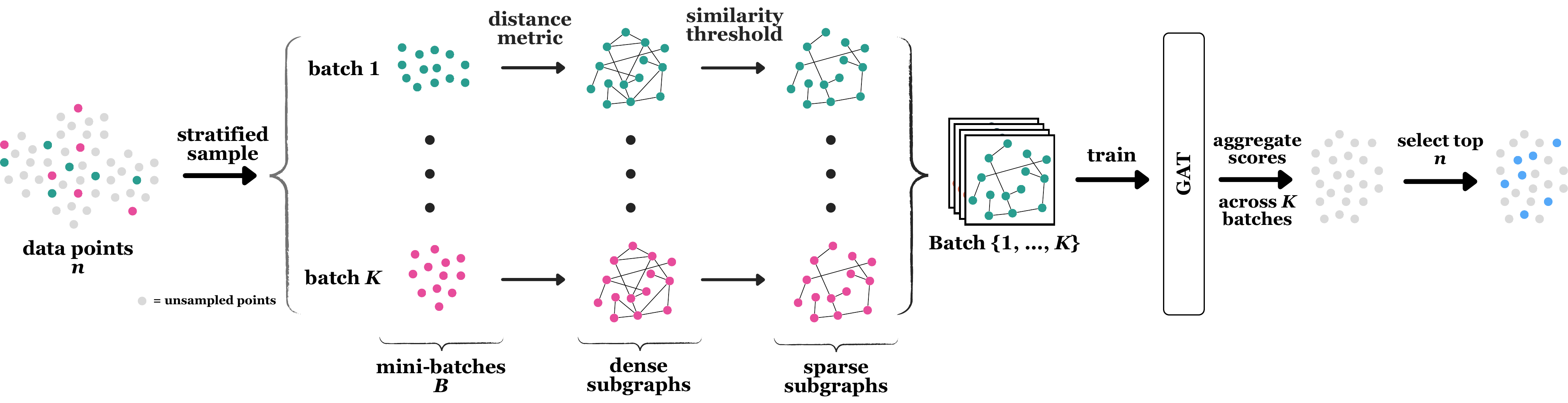}
    \caption{High-level overview of the mini-batch distance metric-based (DM) GAIS. The method samples $K$ mini-batches from the dataset $\mathcal{D}$ without replacement while preserving class distributions and ensuring each batch contains at most $w_{\max}$ instances. For each mini-batch, we compute pairwise distances to construct dense subgraphs, which are then sparsified using similarity thresholds to retain only the strongest connections. The resulting subgraphs are stacked and processed by a graph attention network (GAT). Lastly, attention weights from the trained GAT are aggregated across all $K$ mini-batches to compute node importance scores for selection.}
    \label{fig:methodology_mini_batch}
\end{figure*}

The distance-based method constructs a graph by computing pairwise distances between data points. To reduce computational overhead, we introduce an adaptive mini-batch approach that partitions the dataset into manageable subsets (\textit{mini-batches}) and constructs local graphs within them. The overall process is depicted in Figure~\ref{fig:methodology_mini_batch}.

Given a dataset \(\mathcal{D} = \{(x_i, y_i)\}_{i=1}^n\), where \(x_i \in \mathbb{R}^d\) and \(y_i\) is the corresponding class label, we sample \(K\) mini-batches \(\{B_1, \ldots, B_K\}\) from \(\mathcal{D}\) using stratified sampling without replacement to preserve class distribution.
Computing pairwise distances within each \(B_k\) reduces the overall complexity from \(O(n^2)\) to \(O(K \times w_{\max}^2)\), where \(w_{\max}\) is the maximum instances per mini-batch. The choice of \(K\) balances efficiency and model performance; larger \(K\) offers finer processing but increases overhead, while \(w_{\max}\) provides computational bounds.

For each mini-batch \(B_k\), we employ a stratified sampling approach with a dynamic sampling fraction \(f_k\):
\[
f_k = \min\left(\frac{w_{\text{max}}}{n},\, f_{\text{original}}\right),
\]
where \(f_{\text{original}}\) is the original sampling fraction. This ensures \(|B_k| \leq w_{\max}\) by reducing the sampling fraction whenever \(f_{\text{original}} \cdot n\) would exceed \(w_{\max}\). It also preserves the class distribution \(P(y = c \,\mid\, x \in B_k) \approx P(y = c \,\mid\, x \in \mathcal{D})\) for all classes~\(c\).
This constraint-based sampling approach differs from traditional partitioning by implementing strategic sampling with computational bounds. Rather than exhaustively dividing the dataset into $K$ equal parts, we sample $K$ mini-batches where each contains at most \(w_{\max}\) instances. This ensures scalable complexity regardless of dataset size.

The sampling process maintains a set \(\mathcal{S}_k\) of sampled instances, updated after each mini-batch: $\mathcal{S}_k = \mathcal{S}_{k-1} \cup B_k$ with $\quad \mathcal{S}_0 = \emptyset.$
Subsequent mini-batches are sampled from \(\mathcal{D}_k = \mathcal{D} \setminus \mathcal{S}_{k-1}\). Let \(C_c = \{(x_i, y_i) \in \mathcal{D} : y_i = c\}\) denote the set of all instances from class~\(c\). To ensure each class is represented in every mini-batch, we enforce
$\min_{c}\,\left|\,B_k \cap C_c\right| \;\geq\; 1.$

For each mini-batch $B_k$, we construct a graph $G_k = (V_k, E_k)$, where $V_k = B_k$ and edges $E_k$ are determined by computing a similarity matrix $S_k$ for all pairs $(u, v)$ in $B_k$:
\[
S_{k,uv} = \begin{cases}
    1 - d(x_u, x_v) & \text{if } d \text{ is cosine distance} \\
    1 / (1 + d(x_u, x_v)) & \text{otherwise}
\end{cases}
\]
with \(d(x_u, x_v)\) denoting a chosen distance metric (e.g., Euclidean). To control graph sparsity, we introduce an adaptive similarity threshold \(\tau_k\) computed as a fixed percentile \(p\) (e.g., the 95th percentile) of the pairwise similarity values:
\[
\tau_k = \mathrm{percentile}\,\left(\{\,S_{k,uv} : u,v \in B_k, u \neq v \},\ p\right).
\]
Edges are then defined as
$E_k = \{(u, v) : S_{k,uv} > \tau_k\},$
thereby retaining only the most similar instance pairs in each mini-batch and achieving consistent sparsity. The final graph \(G = (V, E)\) is constructed by combining all mini-batch graphs:
\[
V = \bigcup_{k=1}^K V_k, \quad E = \bigcup_{k=1}^K E_k.
\]

Computationally, this approach reduces the naive \(O(n^2)\) distance-computation cost to
$O\left(\sum_{k=1}^K |B_k|^2\right) = O(K \times w_{\max}^2)$, since \(|B_k| \leq w_{\max}\) for all mini-batches. This constraint ensures computational complexity remains independent of dataset size, making the approach truly scalable for large datasets. This methodology leverages stochastic approximation principles and stratified random sampling, creating mini-batches that reflect the dataset's underlying distribution and leading to more reliable, generalizable results while keeping computations feasible.
The \(w_{\max}\) constraint is empirically justified through plateau analysis (see \nameref{subsec:component_analysis}), where performance stabilizes at mini-batch sizes well within typical computational limits.

\subsubsection{Locality-Sensitive Hashing-based Method}\label{subsubsec:lsh_based_method}
While the distance-based approach directly computes pairwise similarities, we propose an alternative methodology using Locality-Sensitive Hashing (LSH) to approximate similarities via random projections. This avoids the $O(n^2d)$ complexity of exact distance computation by mapping data points into hash buckets in $O(ndk)$ time per hash table, where $n$ is the number of instances, $d$ is the input dimensionality, and $k$ is the number of projection vectors. Our LSH-based approach encompasses three variants of increasing structural richness: single-level LSH (SL-LSH), multi-level LSH (ML-LSH), and multi-view multi-level LSH (MVML-LSH). Each variant offers different trade-offs between computational efficiency and the granularity or diversity of captured similarities.

\begin{figure*}[!t]
    \centering
    \includegraphics[width=0.8\textwidth]{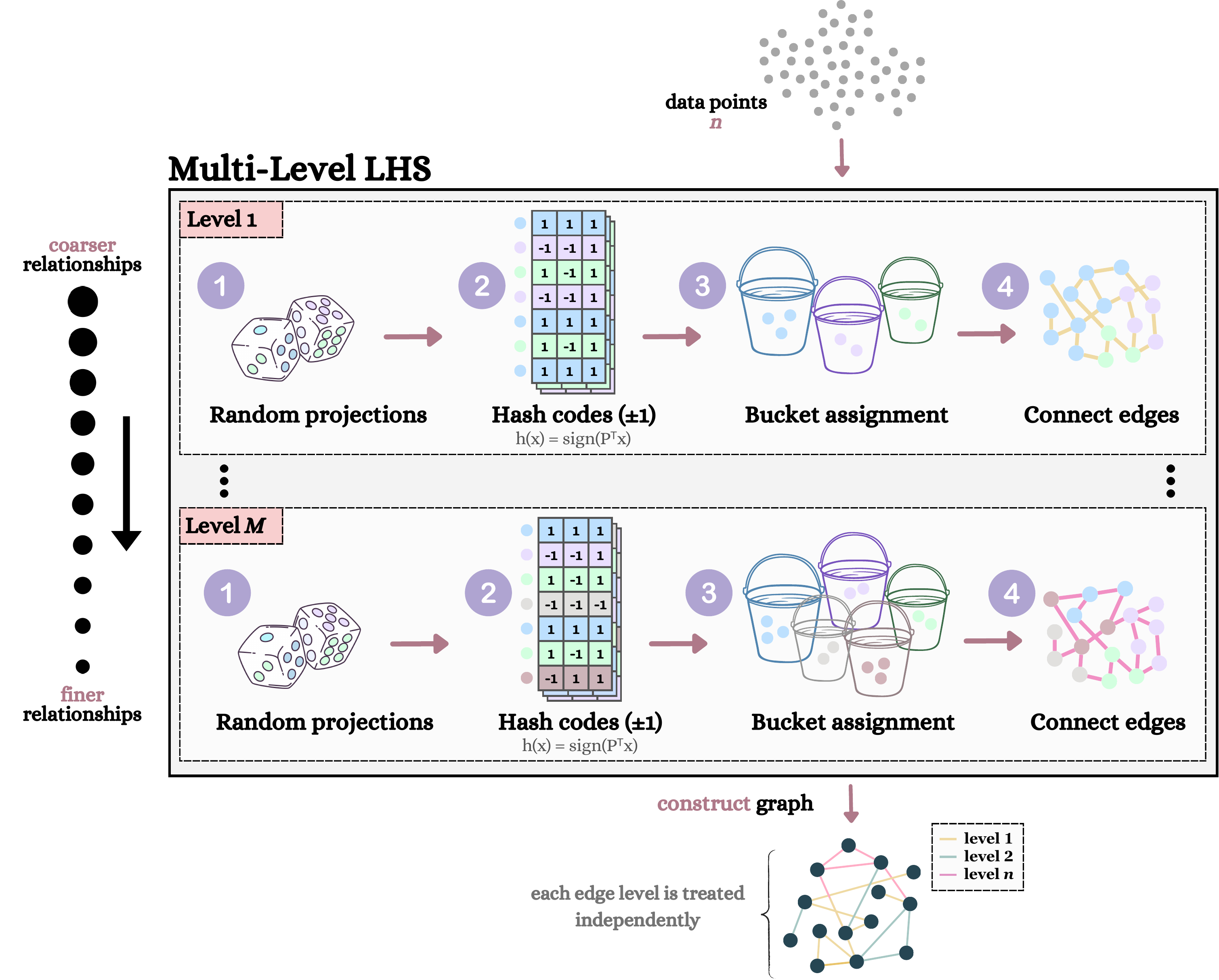}
    \caption{High-level overview of the Multi-Level LSH (ML-LSH) approach. ML-LSH processes instances through multiple hierarchical levels (Level $1$ to $M$), where each level follows a four-step process: \textbf{(1)} Random projections using hash functions, \textbf{(2)} Hash code generation (e.g., $h(x)=\text{sign}(P^Tx)$), \textbf{(3)} Bucket assignment grouping similar instances, and \textbf{(4)} Edge connection within buckets. Lower levels (Level $1$) capture coarse relationships using fewer hash tables and projections, while higher levels (Level $M$) employ exponentially increasing hashing parameters to refine local similarities. The final graph combines edge sets from all levels, with each level treated independently to capture different relationship granularities.}
    \label{fig:methodology_lsh}
\end{figure*}

\paragraph{Single-Level LSH}
Let $\mathcal{D} = \{(x_i, y_i)\}_{i=1}^n$ be a dataset where $x_i \in \mathbb{R}^d$. SL-LSH constructs a single graph by using $L$ independent hash tables, each defined by a random projection matrix $P_l \in \mathbb{R}^{d \times k}$ whose entries are sampled from a standard normal distribution and normalized:
\[
P_l^{(i,j)} \sim \mathcal{N}(0,1)/\sqrt{k}.
\]
For each table $l$, the angular (sign-based) hash function is
\[
h_l(x) \;=\; \mathrm{sign}\left(P_l^{\top} x\right),
\]
while the Euclidean variant introduces a quantization parameter $w$ and random shifts $b_l \in \mathbb{R}^k$:
\[
h_l(x) \;=\; \left\lfloor \frac{P_l^{\top} x + b_l}{\,w\,}\right\rfloor.
\]

Bucket sizes vary with data distribution, so when any bucket \(B_h\) grows larger than a threshold \(\theta\), we perform \emph{adaptive splitting}. Let \(H(x)\) denote the current hash code of a point \(x\). We introduce a new random projection \(p_{\mathrm{new}} \sim \mathcal{N}(0, I_d) / \sqrt{d}\) and extend the hash code to
\[
\tilde{H}(x) \;=\; \left(H(x),\; \mathrm{sign}\left(p_{\mathrm{new}}^{\top}x\right)\right).
\]
All points in \(B_h\) are thus reallocated into finer sub-buckets based on \(\tilde{H}(x)\). This process recurses if any sub-bucket remains above \(\theta\). Conversely, if two buckets \(B_{h_1}, B_{h_2}\) each fall below a merging threshold \(\gamma\) and their Hamming distance is within one bit, we unify them:
\[
\mathrm{Ham}\left(h_1,\; h_2\right) \;\le\; 1,
\quad
\lvert B_{h_1}\rvert + \lvert B_{h_2}\rvert \;<\;\gamma.
\]
Motivated by \citet{Ding2022}, these adaptive strategies balance partitions: dense neighborhoods refine into sub-buckets to capture fine-grained similarities, while sparse regions merge to avoid fragmentation and maintain efficiency.

Finally, we form the graph $G = (V,E)$ by connecting instances $i$ and $j$ that fall into the same bucket in at least one hash table:
\[
E \;=\; \{(i,j) \;\mid\; \exists\, l \in [L] \;\; \mathrm{s.t.}\;\; h_l(x_i) \;=\; h_l(x_j)\}.
\]
This single-level procedure approximates similarity more efficiently than exhaustive pairwise distance calculations, while still identifying closely related data points via common hash signatures.

\paragraph{Multi-Level LSH}
Multi-Level LSH (ML-LSH) builds on the single-level scheme by creating a hierarchy of levels $\{1, \dots, M\}$. At each level $m$, the number of hash tables and projection vectors increases exponentially, and the bucket thresholds decrease:
\[
L_m \;=\; L \cdot 2^m,\quad 
k_m \;=\; k \cdot 2^m,\quad
\theta_m \;=\; \frac{\theta}{\,2^m}.
\]
Lower levels ($m=1$) have fewer tables and looser thresholds that capture broader neighborhoods, whereas higher levels ($m=M$) use more hash tables and finer partitioning. This yields multiple edge sets $\{E_1, \dots, E_M\}$, each reflecting a different “resolution” of similarity:
\[
G_M \;=\; \left(V,\;\{E_1, \dots, E_M\}\right).
\]
By capturing multiple scales of locality, ML-LSH reduces false positives at coarse levels and false negatives at finer levels, thus providing a more nuanced picture of relationships among data points. A high-level schematic of this approach is shown in Figure~\ref{fig:methodology_lsh}.

\paragraph{Multi-View Multi-Level LSH}
Multi-View Multi-Level LSH (MVML-LSH) extends ML-LSH by incorporating $V$ \emph{views}, each with its own hashing parameters and potentially distinct projection schemes (e.g., angular in one view, Euclidean in another). For each view $v \in [V]$ and level $m \in [M]$, the parameters evolve similarly to the multi-level case:
\[
L_{v,m} \;=\; L_v \cdot 2^m,\quad
k_{v,m} \;=\; k_v \cdot 2^m,\quad
\theta_{v,m} \;=\; \frac{\theta_v}{\,2^m}.
\]
Each view $v$ thus produces edge sets $\{E_{v,1}, \dots, E_{v,M}\}$ capturing similarity under a specific metric or hashing scheme. Stacking these across all views yields a heterogeneous graph
\[
G_{V,M} \;=\; \left(V,\;\left\{E_{v,m}\right\}_{v=1,m=1}^{V,M}\right),
\]
where each view-level edge set represents a distinct perspective on data similarity. This multi-view framework allows for complementary insights across different distance metrics or normalization strategies and sets the foundation for cross-view attention mechanisms in subsequent learning stages.

\paragraph{Complexity and Motivations}
Constructing the base LSH tables requires $O(ndk)$ work per table, which is significantly smaller than $O(n^2d)$ for exact pairwise distance computations. Splitting overpopulated buckets and merging sparse ones introduce overhead but do not change the asymptotic complexity. The multi-level hierarchy scales parameters exponentially with level $m$, capturing finer-grained relationships at higher levels, whereas multi-view hashing exploits varied distance notions to uncover complementary similarity structures. This layered design provides a rich, yet computationally feasible, graph representation of the data, facilitating more effective IS while avoiding the prohibitive cost of exact distance-based methods.

\subsection{Graph Attention Network (GAT)}\label{sec:gat-methodology}

GATs provide an ideal foundation for IS due to their ability to learn adaptive node relationships through attention mechanisms. While traditional GATs excel at capturing local structural patterns, IS requires a more nuanced understanding of both local and global data relationships. We build upon the foundational GAT architecture \citep{Velickovic2017} by introducing four key enhancements: (1) head-wise attention amplification through learnable scaling factors, (2) diversity-promoting attention that encourages distinct patterns across attention heads, (3) multi-level/multi-view attention mechanisms that process our graph representations at multiple scales, and (4) cross-level attention via Nystr\"{o}m approximation for efficient global information exchange.

\subsubsection{Standard GAT Formulation}
Let \( G = (V, E)\) be a graph with \(|V|\) nodes and \(|E|\) edges. Each node \(i\in V\) has a feature vector \(\mathbf{x}_i \in \mathbb{R}^d\). A standard GAT layer transforms the node features \(\mathbf{x}_i\) into updated node embeddings \(\mathbf{h}_i\) by computing attention coefficients for each edge \((i,j)\). Specifically, we first apply a linear transformation:
\[
\mathbf{z}_i = W \,\mathbf{x}_i, 
\quad
\mathbf{z}_j = W \,\mathbf{x}_j,
\]
where \(W \in \mathbb{R}^{d' \times d}\) is a learnable weight matrix. We then compute unnormalized attention logits:
\[
e_{ij} = \text{LeakyReLU} \left( \mathbf{a}^\top \left[ \mathbf{z}_i \parallel \mathbf{z}_j \right] \right),
\]
where $\parallel$ denotes concatenation of $\mathbf{z}_i$ and $\mathbf{z}_j$. To get normalized attention coefficients \(\alpha_{ij}\), we apply the softmax over neighbors \(\mathcal{N}_i\):
\[
\alpha_{ij} = \frac{\exp(e_{ij})}{\sum_{k \in \mathcal{N}_i} \exp(e_{ik})}.
\]
Finally, the updated node embedding \(\mathbf{h}_i\) is given by an attention-weighted sum of its neighbors' features:
\[
\mathbf{h}_i = \sigma \!\left(\sum_{j \in \mathcal{N}_i} \alpha_{ij}\,\mathbf{z}_j \right),
\]
where \(\sigma\) is a non-linear activation (i.e.,\ ELU for our case).
To improve stability and capture diverse relational subspaces, we use \(K\) attention heads:
\[
\mathbf{h}_i 
= \big\Vert_{k=1}^K 
\;\sigma \!\left(\sum_{j \in \mathcal{N}_i} \alpha_{ij}^k\,W^k \mathbf{x}_j \right),
\]
where each head \(k\) uses its own linear parameters \(W^k\) and attention coefficients \(\alpha_{ij}^k\).

\subsubsection{Head-wise Attention Amplification}

Multiple attention heads can capture diverse relationship patterns in the input graph. However, for IS, the relative importance of these patterns may vary significantly. We propose a learnable head-wise scaling mechanism that enables the model to automatically weight the contribution of each attention head. For each head \( k \), we maintain a scalar parameter \( \gamma_k \) that globally modulates the attention weights. After computing standard attention coefficients \( \alpha_{ij}^k \), we apply the transformation:
\[
\hat{\alpha}_{ij}^k = (1 + \sigma(\gamma_k))\alpha_{ij}^k
\]
where \( \sigma(\cdot) \) is the sigmoid function. This scaling factor, bounded in \( [1,2] \), allows head \( k \) to amplify its attention pattern if beneficial for the task. The scaled weights are then renormalized within each neighborhood:
\[
\tilde{\alpha}_{ij}^k = \frac{\hat{\alpha}_{ij}^k}{\sum_{l \in \mathcal{N}_i} \hat{\alpha}_{il}^k}
\]
This mechanism serves two key purposes: (1) it allows the model to emphasize attention heads that identify more discriminative relationships, and (2) it provides interpretable weights indicating which types of node relationships are most informative. 

\subsubsection{Diversity-Promoting Attention}
In multi-head attention mechanisms, heads can converge to similar attention patterns, reducing their collective expressiveness. We address this by introducing a diversity-promoting mechanism that encourages heads to learn distinct relationship patterns. For each head $k$ and node $i$, we compute the deviation from mean attention:
\[
\bar{\alpha}_i^k 
= 
\frac{1}{|\mathcal{N}_i|}\sum_{j \in \mathcal{N}_i} \alpha_{ij}^k,
\qquad
\delta_{ij}^k 
= 
\frac{\left|\alpha_{ij}^k - \bar{\alpha}_i^k \right|}
     {\bar{\alpha}_i^k + \epsilon}.
\]
where $\epsilon$ is a small positive constant (typically $\epsilon = 10^{-8}$) added for numerical stability to prevent division by zero when the mean attention $\bar{\alpha}_i^k$ approaches zero.
The final attention weights incorporate this diversity term:
\[
\alpha_{ij}^{k,\text{div}} = \alpha_{ij}^k + \lambda_k\delta_{ij}^k
\]
where $\lambda_k$ is a learnable parameter initialized within $[0.5r, r]$ for small positive constant $r$. Larger values of $\lambda_k$ increase the reward for attention patterns that deviate from the neighborhood mean, pushing each head to specialize in distinct relationship structures. This specialization is particularly important for IS, where identifying diverse relationship patterns can better inform the importance of each instance.

\subsubsection{Multi-Level and Multi-View Extensions}

\paragraph{Multi-Level GAT}
We generalize GAT to handle multiple edge sets \(\{E_1,\dots,E_M\}\) in a single graph, each capturing a different similarity or relationship (e.g., multiple distance thresholds). Let \(\mathcal{N}_i^m\) be the neighbors of \(i\) in edge set \(E_m\). We maintain separate attention mechanisms \(\alpha_{ij}^{m,l}\) for each level \(m\) at layer \(l\), and aggregate:
\[
\mathbf{h}_i^{(l)} 
= 
\frac{1}{M} 
\sum_{m=1}^M 
\sigma\!\left(\sum_{j \in \mathcal{N}_i^m} \alpha_{ij}^{m,l}\, W^l \mathbf{h}_j^{(l-1)}\right).
\]
Mean aggregation across levels enables uniform information flow across heterogeneous relationship types.

\paragraph{Multi-View Multi-Level GAT}
For more diverse relationships, we may also have multiple \emph{views}, denoted by \(\tau \in \mathcal{T}\), each containing multiple levels \(\{E_{\tau,1}, \dots, E_{\tau,M}\}\). We can maintain separate transformations \(W_\tau^l\) and attention \(\alpha_{ij}^{\tau,m,l}\) for each view \(\tau\) and each level \(m\):
\[
\mathbf{h}_i^{(l)} 
= 
\frac{1}{|\mathcal{T}|} 
\sum_{\tau \in \mathcal{T}} 
\sigma \!\left(
  \sum_{m=1}^M 
    \sum_{j \in \mathcal{N}_i^{\tau,m}} 
      \alpha_{ij}^{\tau,m,l}\,
      W_\tau^l\,
      \mathbf{h}_j^{(l-1)}
\right).
\]
This multi-view extension allows the model to integrate information from different similarity metrics (views $\tau$), while maintaining the benefits of multi-level aggregation within each view.

\subsubsection{Cross-Level Attention via Nystr\"{o}m Approximation}
Although combining \emph{all} levels/views can be powerful, direct attention across all nodes at all levels can become prohibitively expensive.
To enable scalable \textit{cross-level} attention, we adapt the Nystr\"{o}m method \citep{Xiong2021} to approximate the global attention matrix. Let \(\{\mathbf{h}_i^{(1)}, \dots, \mathbf{h}_i^{(L)}\}\) be node \(i\)'s embeddings from \(L\) different levels. The naive approach of attending to all nodes across all levels would incur \(O(n^2LH)\) complexity, where \(H\) is the number of attention heads.
In the Nystr\"{o}m approximation, landmark nodes serve as representative anchors that create a low-rank approximation of the full attention matrix. By computing attention only between all nodes and this smaller set of landmarks (rather than between all pairs of nodes), we can substantially reduce computational requirements while preserving the essential structure of global attention.
We reduce this by selecting \(m \ll n\) landmark nodes through a hybrid sampling strategy:
\[
\mathcal{L} = \mathcal{L}_{\text{deg}} \cup \mathcal{L}_{\text{rand}}, \quad |\mathcal{L}_{\text{deg}}| = \lceil\gamma m\rceil, \quad |\mathcal{L}_{\text{rand}}| = m - |\mathcal{L}_{\text{deg}}|
\]
where \(\gamma\) is the degree ratio parameter and \(\mathcal{L}_{\text{deg}}\) contains the nodes with highest total degrees.
Using these landmarks, the Nystr\"{o}m approximation factorizes the attention computation as:
\[
\tilde{\mathbf{h}}_i = \text{softmax}\left(\frac{Q_i K_{\mathcal{L}}^T}{\sqrt{d}}\right)V_{\mathcal{L}}
\]
where \(Q_i, K_{\mathcal{L}}, V_{\mathcal{L}}\) are the query, key, and value projections respectively, and \(d\) is the head dimension. This reduces the complexity to \(O(nmLH)\).

The cross-attended features are combined with the original embeddings through a learned gating mechanism:
\[
g_i \;=\; \sigma\!\left(W_g\,[\,\mathbf{h}_i \,\|\; \tilde{\mathbf{h}}_i]\right),
\qquad
\mathbf{h}_i^{\mathrm{final}} 
\,=\, 
g_i\,\mathbf{h}_i 
\;+\; 
\left(1 - g_i\right)\,\tilde{\mathbf{h}}_i.
\]
This adaptive combination preserves useful local structures while incorporating global cross-level context when beneficial.

The effectiveness of Nystr\"{o}m-based approach for IS stems from two key properties: (1) the landmark selection strategy ensures coverage of both high-degree nodes and a random sample, providing a representative basis for attention computation across different graph regions, and (2) the gating mechanism allows adaptive interpolation between local and cross-level representations, critical for identifying instances with varying structural roles in the graph.


\subsection{Instance Selection}
The final component of our framework involves computing node importance scores from the learned attention weights and employing these scores for selective instance sampling.

\subsubsection{Node Importance Computation}
We propose a \emph{dual-perspective} importance scoring mechanism that considers both the sender (outgoing attention) and receiver (incoming attention) roles of nodes in the attention graph. The final node importance score is computed as:
\[
I(i) = w_s \cdot I_s(i) + w_r \cdot I_r(i)
\]
where the sender importance \( I_s(i) \) captures how influential a node is in providing information:
\[
I_s(i) = \frac{1}{\sqrt{d_{\text{out}}(i) + 1}} \sum_{l=1}^L \tilde{\beta}_l \sum_{j \in \mathcal{N}_{\text{out}}(i)} \sum_{k=1}^K \alpha_{ij}^{(l,k)}
\]
and the receiver importance \( I_r(i) \) measures information aggregation significance:
\[
I_r(i) = \frac{1}{\sqrt{d_{\text{in}}(i) + 1}} \sum_{l=1}^L \tilde{\beta}_l \sum_{j \in \mathcal{N}_{\text{in}}(i)} \sum_{k=1}^K \alpha_{ji}^{(l,k)}
\]
Here, \( \tilde{\beta}_l = \beta_l / \sum_{m=1}^L \beta_m \) are normalized layer weights allowing controlled influence of deeper layers, \( \alpha_{ij}^{(l,k)} \) represents the attention weight from node \( i \) to \( j \) in layer \( l \) and head \( k \). The degree normalization terms \( d_{\text{in}}(i) \) and \( d_{\text{out}}(i) \) prevent high-degree node dominance while preserving the structural significance of well-connected instances.

\subsubsection{Selection Strategies}
We introduce two complementary approaches for selecting instances: global threshold-based and class-aware selection. Given a desired retention rate \(\rho\in(0,1)\), the global threshold approach selects the top \( \rho N \) instances based on their importance scores:
\[
\mathcal{S}_{\text{global}} = \text{top}_{\rho N}(\{i \mid i \in \mathcal{V}\})
\]

The class-aware selection strategies address potential class imbalance through two variants. The balanced strategy ensures equal representation across classes:
\[
\mathcal{S}_{\text{balanced}} = \bigcup_{c \in \mathcal{C}} \text{top}_{\lfloor \rho N/|\mathcal{C}| \rfloor}(\{i \mid y_i = c\})
\]
while the proportional strategy maintains the original class distribution:
\[
\mathcal{S}_{\text{proportional}} = \bigcup_{c \in \mathcal{C}} \text{top}_{\lfloor \rho N p_c \rfloor}(\{i \mid y_i = c\})
\]
where \( p_c \) represents the original class proportion. Both strategies select instances within each class based on their importance scores, ensuring that the structural significance captured by the attention mechanism is preserved while maintaining desired class distributions.


\begin{table*}[!ht]
\centering
\caption{Dataset statistics with domain, number of instances ($N$), features ($D$), and classes ($K$). Dataset size indicators: $^{\dagger}$small ($N < 10,000$), $^{\ddagger}$medium ($10,000 \leq N \leq 100,000$), $^{\star}$large ($N > 100,000$).}
\label{tab:dataset_statistics}
{\scriptsize
\begin{tabular}{llrrr|llrrr}
\toprule
Dataset & Domain & N & D & K & Dataset & Domain & N & D & K \\
\midrule
heart$^{\dagger}$ & Healthcare & 303 & 13 & 2 & thyroid$^{\dagger}$ & Healthcare & 7,200 & 21 & 3 \\
hcc$^{\dagger}$ & Healthcare & 615 & 12 & 2 & ringnorm$^{\dagger}$ & Synthetic Data & 7,400 & 20 & 2 \\
stroke$^{\dagger}$ & Healthcare & 749 & 10 & 2 & twonorm$^{\dagger}$ & Synthetic Data & 7,400 & 20 & 2 \\
diabetes$^{\dagger}$ & Healthcare & 768 & 8 & 2 & coil2000$^{\dagger}$ & Computer Vision & 9,822 & 85 & 2 \\
german$^{\dagger}$ & Finance & 1,000 & 20 & 2 & pen$^{\ddagger}$ & Computer Vision & 10,992 & 16 & 10 \\
contraceptive$^{\dagger}$ & Healthcare & 1,473 & 9 & 3 & nursery$^{\ddagger}$ & Social Science & 12,958 & 8 & 4 \\
yeast$^{\dagger}$ & Bioinformatics & 1,479 & 8 & 9 & dry\_bean$^{\ddagger}$ & Biology & 13,611 & 16 & 7 \\
car$^{\dagger}$ & Automotive & 1,728 & 6 & 4 & magic$^{\ddagger}$ & Physics & 19,020 & 10 & 2 \\
titanic$^{\dagger}$ & Social Science & 2,201 & 3 & 2 & letter$^{\ddagger}$ & Computer Vision & 20,000 & 16 & 26 \\
segment$^{\dagger}$ & Computer Vision & 2,310 & 19 & 7 & adult$^{\ddagger}$ & Social Science & 32,561 & 12 & 2 \\
splice$^{\dagger}$ & Bioinformatics & 3,190 & 61 & 2 & shuttle$^{\ddagger}$ & Physics & 58,000 & 7 & 7 \\
chess$^{\dagger}$ & Game Theory & 3,196 & 36 & 2 & mnist$^{\ddagger}$ & Computer Vision & 70,000 & 784 & 10 \\
abalone$^{\dagger}$ & Biology & 4,168 & 8 & 21 & fars$^{\star}$ & Transportation & 100,968 & 29 & 8 \\
spambase$^{\dagger}$ & Text Classification & 4,601 & 57 & 2 & ldpa$^{\star}$ & Healthcare & 164,860 & 7 & 11 \\
banana$^{\dagger}$ & Synthetic Data & 5,300 & 2 & 2 & census-income$^{\star}$ & Social Science & 199,523 & 41 & 2 \\
phoneme$^{\dagger}$ & Speech Recognition & 5,404 & 5 & 2 & skin$^{\star}$ & Computer Vision & 245,057 & 3 & 2 \\
page\_blocks$^{\dagger}$ & Document Analysis & 5,473 & 10 & 5 & covertype$^{\star}$ & Environmental Science & 581,012 & 54 & 7 \\
texture$^{\dagger}$ & Computer Vision & 5,500 & 40 & 11 & pokerhand$^{\star}$ & Game Theory & 1,025,010 & 10 & 10 \\
opt-digits$^{\dagger}$ & Computer Vision & 5,620 & 64 & 10 & susy$^{\star}$ & Physics & 5,000,000 & 18 & 2 \\
satellite$^{\dagger}$ & Remote Sensing & 6,435 & 36 & 6 & & & & & \\
\midrule
\multicolumn{10}{c}{\textbf{Average:}\hspace{0.5em}$N$ = \textbf{196,075},\hspace{0.5em}$D$ = \textbf{41},\hspace{0.5em}$K$ = \textbf{6}} \\
\bottomrule
\end{tabular}
}
\end{table*}

\section{Experimental Setup}\label{sec:experimental_setup}

\subsection{Datasets and Models}
This study employs diverse classification datasets from the UCI Machine Learning Repository\footnote{UCI Machine Learning Repository. Kelly, M., Longjohn, R., Nottingham, K. (2023). Available at: \url{https://archive.ics.uci.edu} (accessed 4 October 2024).} to evaluate our proposed GAIS method. Table \ref{tab:dataset_statistics} summarizes these datasets, including their domain, number of instances ($N$), features ($D$), and classes ($K$). The datasets vary widely in size (from 303 to five million instances), dimensionality (two to 784 features), and class distributions (balanced and imbalanced), spanning domains such as healthcare, game theory, and bioinformatics.
We evaluate our method using Random Forest and Logistic Regression as representative non-linear and linear classifiers, respectively.

\subsection{Data Preprocessing}

We perform stratified sampling to split each dataset into training (80\%), validation (10\%), and test (10\%) sets, maintaining class distributions. The training set is used for graph construction, GAT model training, and IS; the validation set for hyperparameter tuning; and test set for final evaluation.
All features are retained, with categorical variables label-encoded and numerical features normalized to [0, 1] via min-max scaling.

\subsection{Implementation Details}

Experiments were conducted on a Windows 10 machine with a 12th Gen Intel Core i7, 128\,GB RAM, and an NVIDIA GeForce RTX 4090 GPU (24\,GB VRAM), using Python 3.10, PyTorch 2.4.0, and PyTorch Geometric 2.5.3. For reproducibility, we used a single core where applicable and a consistent random seed across processes, while preserving the inherent randomness of stochastic methods.

Our GAT implementation employs a multi-layer architecture with training limited to 500 iterations and early stopping (patience of 30). We use the Adam optimizer (learning rate: $5 \times 10^{-3}$, weight decay: $5 \times 10^{-4}$) and a plateau-based learning rate scheduler (reduction factor: 0.75, patience: 25, min\_lr: $10^{-5}$). The model is trained using cross-entropy loss.

\subsection{Hyperparameter Optimization}
\begin{figure}[!t]
    \centering
    \includegraphics[width=0.7\linewidth]{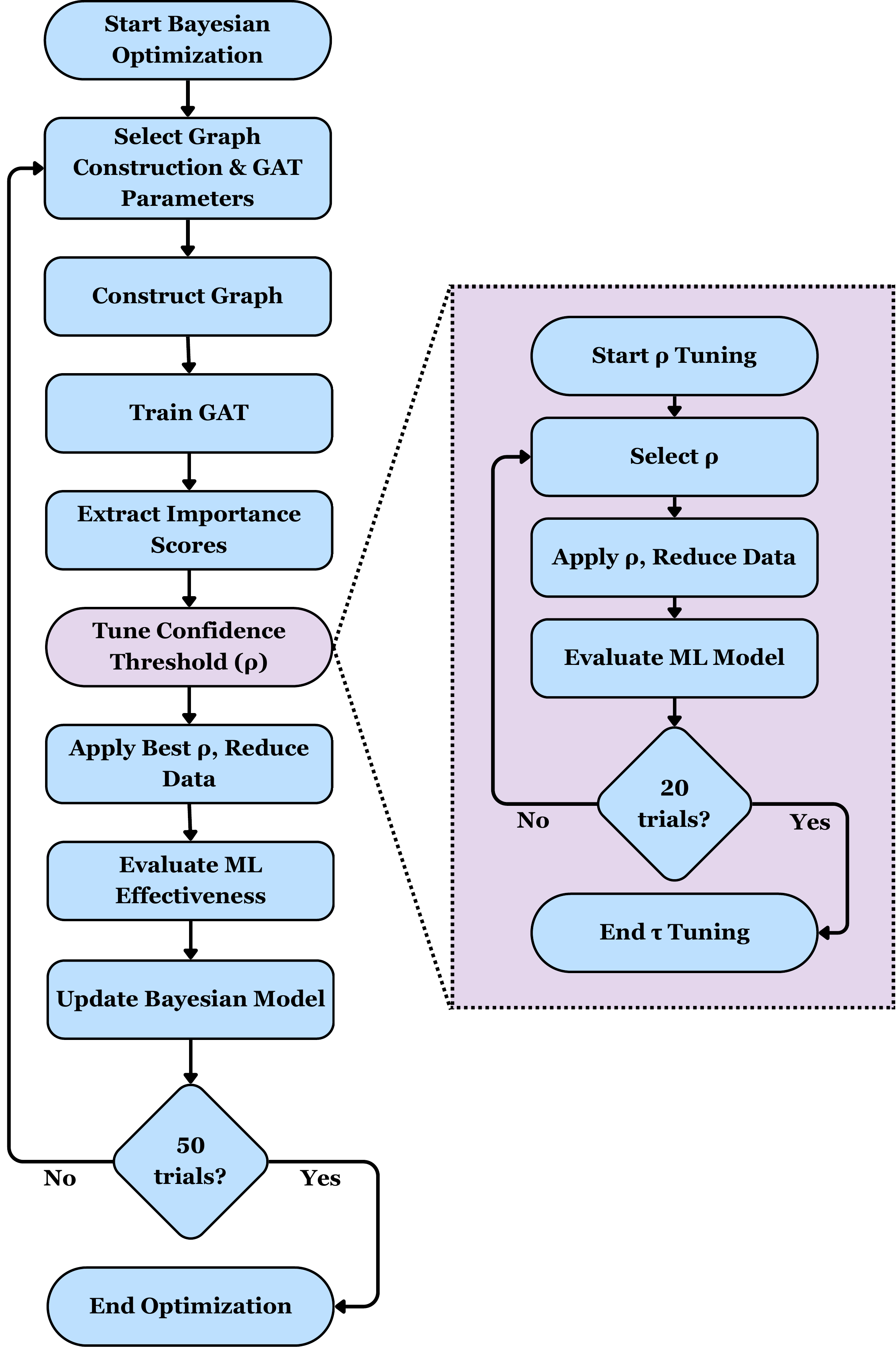}
    \caption{Hyperparameter tuning procedure using Bayesian optimization with nested trials. The main loop (left) conducts 50 trials optimizing graph construction and GAT parameters, while each trial includes a nested $\rho$-tuning subprocess (right) that performs 20 trials to optimize instance selection retention rates and strategies.}
    \label{fig:tuning_flowchart}
\end{figure}

We employed Bayesian optimization using the Optuna framework \citep{Akiba2019} for hyperparameter tuning across all components of our framework. For each dataset and method, we conduct 50 primary optimization trials to tune the graph construction parameters and GAT architectural components. Within each primary trial, we perform nested optimization of the IS process through 20 additional trials. This nested optimization determines both the selection strategy (global threshold-based or class-aware) and its associated parameters, including retention rates ($\rho$). This hierarchical optimization approach ensures that the selection strategy is tailored to the specific characteristics of the graph representation and importance scores produced by each configuration.
The complete list of tuned hyperparameters and their ranges is provided in the supplementary material.
The overall procedure is depicted in Figure~\ref{fig:tuning_flowchart}.

\subsection{Evaluation Metrics}\label{subsec:evaluation_metrics}

Table \ref{tab:evaluation_metrics} summarizes our evaluation metrics, combining standard classification performance measures with instance reduction metrics to assess both predictive capability and data efficiency \citep{Malhat2020}, along with computational time measurements to evaluate practical scalability.

\begin{table}[h]
\centering
\caption{Metrics used in GAIS evaluation. Classification metrics use standard definitions where TP=True Positives, TN=True Negatives, FP=False Positives, FN=False Negatives. For multi-class problems, precision, recall, and F1 are macro-averaged.}
\label{tab:evaluation_metrics}
\scriptsize
\begin{tabular*}{\linewidth}{@{\extracolsep\fill}
    >{\raggedright\arraybackslash}p{0.29\linewidth}
    >{\raggedright\arraybackslash}p{0.67\linewidth}}
\toprule
Metric & Formula / Description \\
\midrule
Accuracy & $AC$ = (TP + TN) / (TP + TN + FP + FN) \\
Precision & $PR$ = TP / (TP + FP) \\
Recall & $RE$ = TP / (TP + FN) \\
F1 Score & $F1$ = 2 · ($PR \times RE$) / ($PR + RE$) \\
Reduction Rate & $R$ = 1 - (\text{Selected Instances} / \text{Total Instances}) \\
Effectiveness & $E$ = $AC \times R$  \\
Graph Const. Time & Time to build graph representation \\
GAT Training Time & Time to train graph attention network \\
IS Time & Time for selecting instances \\
ML Training Time & Time to train ML model \\
\bottomrule
\end{tabular*}
\end{table}

\section{Results and Discussions}\label{sec:results}

This study proposes a graph-based instance selection algorithm (GAIS) utilizing GATs to identify informative instances. This section presents and discusses our experimental findings across multiple variants of GAIS using different graph construction methods: (1) distance-based mini-batching (DM) and (2) locality-sensitive hashing (LSH) based approaches, including single-level (SL-LSH), multi-level (ML-LSH), and multi-view multi-level (MVML-LSH) variants. All reported results represent the mean performance across five independent runs with a fixed random seed to ensure reproducibility.

\begin{table*}[!t]
\caption{Performance comparison of GAIS variants across dataset sizes. Methods include distance-based mini-batching (DM), locality-sensitive hashing variants (SL/ML/MVML-LSH with cross-attention C), and random sampling baselines. Values show mean $\pm$ standard deviation across multiple runs. Bold indicates best performance per dataset category.}
\label{table:summary_performance_metrics}
\scriptsize
\begin{tabular*}{\textwidth}{@{\extracolsep\fill}
    >{\raggedright\arraybackslash}p{0.11\textwidth}|
    >{\centering\arraybackslash}p{0.09\textwidth}
    >{\centering\arraybackslash}p{0.09\textwidth}
    >{\centering\arraybackslash}p{0.09\textwidth}
    >{\centering\arraybackslash}p{0.09\textwidth}
    >{\centering\arraybackslash}p{0.09\textwidth}
    >{\centering\arraybackslash}p{0.09\textwidth}
    >{\centering\arraybackslash}p{0.09\textwidth}}
\toprule
Method & $AC$ & $PR$ & $RE$ & $F1$ & $E$ & $E_{F1}$ & $R$ \\
\midrule
\multicolumn{8}{c}{\textbf{Small Datasets ($< 10,000$)}} \\
\midrule
Original & 0.860 ± 0.183 & 0.799 ± 0.270 & 0.734 ± 0.299 & 0.758 ± 0.284 & --- & --- & --- \\
Random (1\%) & 0.718 ± 0.194 & 0.588 ± 0.309 & 0.510 ± 0.322 & 0.499 ± 0.298 & 0.711 ± 0.192 & 0.494 ± 0.295 & 0.990 ± 0.000 \\
Random (3\%) & 0.784 ± 0.182 & 0.716 ± 0.271 & 0.585 ± 0.310 & 0.600 ± 0.292 & 0.761 ± 0.177 & 0.582 ± 0.283 & 0.970 ± 0.000 \\
\cmidrule(r){1-1} \cmidrule(r){2-8}
Mini-batch DM & 0.818 ± 0.169 & 0.739 ± 0.249 & 0.693 ± 0.271 & 0.703 ± 0.260 & 0.791 ± 0.168 & 0.678 ± 0.252 & 0.965 ± 0.023 \\
SL-LSH & 0.813 ± 0.169 & 0.705 ± 0.249 & 0.680 ± 0.280 & 0.682 ± 0.260 & 0.786 ± 0.170 & 0.659 ± 0.254 & \textbf{0.966 ± 0.023} \\
ML-LSH & 0.823 ± 0.174 & 0.726 ± 0.257 & 0.675 ± 0.288 & 0.686 ± 0.270 & 0.792 ± 0.173 & 0.659 ± 0.260 & 0.961 ± 0.029 \\
CML-LSH & 0.823 ± 0.170 & 0.720 ± 0.255 & 0.674 ± 0.288 & 0.684 ± 0.268 & 0.793 ± 0.170 & 0.657 ± 0.259 & 0.961 ± 0.025 \\
MVML-LSH & 0.824 ± 0.173 & 0.730 ± 0.256 & 0.682 ± 0.294 & 0.689 ± 0.273 & 0.794 ± 0.171 & 0.662 ± 0.263 & 0.963 ± 0.026 \\
CMVML-LSH & \textbf{0.825 ± 0.171} & \textbf{0.745 ± 0.240} & \textbf{0.696 ± 0.285} & \textbf{0.704 ± 0.264} & \textbf{0.796 ± 0.171} & \textbf{0.678 ± 0.255} & 0.963 ± 0.025 \\
\midrule
\multicolumn{8}{c}{\textbf{Medium Datasets ($10,000 - 100,000$)}} \\
\midrule
Original & 0.945 ± 0.052 & 0.897 ± 0.108 & 0.876 ± 0.144 & 0.884 ± 0.128 & --- & --- & --- \\
Random (1\%) & 0.826 ± 0.116 & 0.748 ± 0.119 & 0.709 ± 0.176 & 0.711 ± 0.161 & 0.818 ± 0.115 & 0.704 ± 0.160 & 0.990 ± 0.000 \\
Random (3\%) & 0.873 ± 0.076 & 0.788 ± 0.118 & 0.767 ± 0.158 & 0.769 ± 0.145 & 0.847 ± 0.074 & 0.746 ± 0.140 & 0.970 ± 0.000 \\
\cmidrule(r){1-1} \cmidrule(r){2-8}
Mini-batch DM & \textbf{0.896 ± 0.055} & 0.812 ± 0.148 & 0.776 ± 0.187 & 0.780 ± 0.170 & \textbf{0.872 ± 0.063} & 0.758 ± 0.162 & \textbf{0.973 ± 0.022} \\
SL-LSH & 0.895 ± 0.051 & \textbf{0.816 ± 0.160} & 0.781 ± 0.181 & 0.783 ± 0.168 & 0.865 ± 0.063 & 0.755 ± 0.160 & 0.966 ± 0.031 \\
ML-LSH & 0.894 ± 0.054 & 0.810 ± 0.148 & \textbf{0.789 ± 0.184} & 0.789 ± 0.166 & 0.866 ± 0.061 & 0.762 ± 0.157 & 0.968 ± 0.024 \\
CML-LSH & \textbf{0.896 ± 0.053} & 0.812 ± 0.146 & 0.787 ± 0.184 & 0.788 ± 0.167 & 0.869 ± 0.062 & 0.762 ± 0.158 & 0.969 ± 0.025 \\
MVML-LSH & 0.895 ± 0.054 & 0.810 ± 0.152 & 0.784 ± 0.183 & \textbf{0.790 ± 0.166} & 0.868 ± 0.062 & \textbf{0.764 ± 0.155} & 0.970 ± 0.021 \\
CMVML-LSH & 0.893 ± 0.057 & 0.809 ± 0.155 & 0.776 ± 0.192 & 0.781 ± 0.171 & 0.864 ± 0.062 & 0.754 ± 0.159 & 0.968 ± 0.020 \\
\midrule
\multicolumn{8}{c}{\textbf{Large Datasets ($> 100,000$)}} \\
\midrule
Original & 0.880 ± 0.092 & 0.782 ± 0.184 & 0.645 ± 0.269 & 0.683 ± 0.259 & --- & --- & --- \\
Random (1\%) & 0.782 ± 0.144 & 0.652 ± 0.251 & 0.483 ± 0.284 & 0.518 ± 0.278 & 0.774 ± 0.143 & 0.513 ± 0.276 & 0.990 ± 0.000 \\
Random (3\%) & 0.808 ± 0.126 & 0.675 ± 0.244 & 0.520 ± 0.271 & 0.559 ± 0.267 & 0.784 ± 0.122 & 0.542 ± 0.259 & 0.970 ± 0.000 \\
\cmidrule(r){1-1} \cmidrule(r){2-8}
Mini-batch DM & 0.810 ± 0.126 & 0.680 ± 0.224 & 0.520 ± 0.280 & 0.554 ± 0.272 & 0.795 ± 0.131 & 0.544 ± 0.272 & \textbf{0.980 ± 0.024} \\
SL-LSH & 0.823 ± 0.111 & 0.685 ± 0.233 & 0.516 ± 0.296 & 0.547 ± 0.291 & 0.799 ± 0.123 & 0.531 ± 0.288 & 0.969 ± 0.032 \\
ML-LSH & 0.824 ± 0.111 & 0.706 ± 0.221 & 0.516 ± 0.299 & 0.548 ± 0.295 & 0.798 ± 0.124 & 0.532 ± 0.291 & 0.966 ± 0.034 \\
CML-LSH & 0.823 ± 0.112 & 0.692 ± 0.234 & 0.523 ± 0.283 & 0.558 ± 0.276 & 0.800 ± 0.123 & 0.544 ± 0.274 & 0.971 ± 0.029 \\
MVML-LSH & \textbf{0.840 ± 0.109} & 0.696 ± 0.208 & 0.477 ± 0.294 & 0.519 ± 0.280 & \textbf{0.819 ± 0.123} & 0.506 ± 0.277 & 0.973 ± 0.031 \\
CMVML-LSH & 0.823 ± 0.114 & \textbf{0.715 ± 0.219} & \textbf{0.528 ± 0.292} & \textbf{0.563 ± 0.287} & 0.800 ± 0.122 & \textbf{0.547 ± 0.283} & 0.970 ± 0.029 \\
\bottomrule
\end{tabular*}
\end{table*}

\begin{table*}
\centering
\caption{Computational time comparison (in seconds) of GAIS variants across dataset sizes. Methods include distance-based mini-batching (DM) and locality-sensitive hashing variants (SL/ML/MVML-LSH with cross-attention C). Time components: graph construction (GCT), GAT training (GTT), instance selection (IST), and machine learning model training on selected instances (MLT). Values show mean $\pm$ standard deviation. Bold indicates fastest performance per dataset category.}
\label{table:summary_timing_metrics}
\scriptsize
\begin{tabular*}{0.65\textwidth}{@{\extracolsep\fill}
    >{\raggedright\arraybackslash}p{0.1\textwidth}|
    >{\centering\arraybackslash}p{0.1\textwidth}
    >{\centering\arraybackslash}p{0.1\textwidth}
    >{\centering\arraybackslash}p{0.1\textwidth}
    >{\centering\arraybackslash}p{0.1\textwidth}}
\toprule
Method & GCT & GTT & IST & MLT \\
\midrule
\multicolumn{5}{c}{\textbf{Small Datasets ($< 10,000$)}} \\
\midrule
Original & --- & --- & --- & 0.31 ± 0.53 \\
Mini-batch DM & \textbf{0.05 ± 0.05} & 3.72 ± 3.44 & 0.00 ± 0.00 & 0.06 ± 0.03 \\
SL-LSH & 0.17 ± 0.15 & \textbf{2.55 ± 2.49} & 0.00 ± 0.00 & 0.09 ± 0.05 \\
ML-LSH & 1.89 ± 6.97 & 4.79 ± 4.01 & 0.00 ± 0.00 & \textbf{0.05 ± 0.03} \\
CML-LSH & 1.96 ± 7.04 & 4.85 ± 3.64 & 0.00 ± 0.00 & 0.06 ± 0.03 \\
MVML-LSH & 1.51 ± 2.11 & 5.94 ± 7.72 & 0.00 ± 0.00 & 0.07 ± 0.05 \\
CMVML-LSH & 0.95 ± 0.68 & 4.04 ± 3.07 & 0.00 ± 0.00 & 0.07 ± 0.04 \\
\midrule
\multicolumn{5}{c}{\textbf{Medium Datasets ($10,000 - 100,000$)}} \\
\midrule
Original & --- & --- & --- & 2.05 ± 1.22 \\
Mini-batch DM & \textbf{0.35 ± 0.35} & \textbf{6.19 ± 8.88} & \textbf{0.00 ± 0.00} & \textbf{0.10 ± 0.04} \\
SL-LSH & 0.90 ± 1.03 & 6.48 ± 2.93 & 0.01 ± 0.02 & 0.18 ± 0.09 \\
ML-LSH & 2.27 ± 2.89 & 16.68 ± 18.65 & 0.02 ± 0.03 & 0.15 ± 0.10 \\
CML-LSH & 2.30 ± 3.19 & 20.18 ± 27.29 & 0.02 ± 0.03 & 0.11 ± 0.05 \\
MVML-LSH & 4.62 ± 4.96 & 13.77 ± 9.49 & 0.01 ± 0.02 & 0.12 ± 0.05 \\
CMVML-LSH & 6.00 ± 7.80 & 33.25 ± 57.33 & 0.02 ± 0.03 & 0.14 ± 0.08 \\
\midrule
\multicolumn{5}{c}{\textbf{Large Datasets ($> 100,000$)}} \\
\midrule
Original & --- & --- & --- & 70.64 ± 158.70 \\
Mini-batch DM & \textbf{2.55 ± 1.78} & \textbf{10.31 ± 10.91} & \textbf{0.01 ± 0.01} & \textbf{0.28 ± 0.32} \\
SL-LSH & 7.46 ± 6.93 & 28.98 ± 32.56 & 0.40 ± 0.80 & 0.82 ± 0.79 \\
ML-LSH & 10.44 ± 10.65 & 43.37 ± 41.89 & 0.29 ± 0.48 & 0.36 ± 0.36 \\
CML-LSH & 15.12 ± 15.81 & 106.05 ± 129.30 & 0.26 ± 0.52 & 0.36 ± 0.46 \\
MVML-LSH & 42.90 ± 48.28 & 80.69 ± 50.17 & 0.26 ± 0.57 & 0.51 ± 0.73 \\
CMVML-LSH & 35.08 ± 38.76 & 103.14 ± 124.79 & 0.26 ± 0.52 & 0.36 ± 0.32 \\
\bottomrule
\end{tabular*}
\end{table*}

\subsection{Comparing GAIS Variants}

Table \ref{table:summary_performance_metrics} presents a summary of performance metrics across GAIS variants, original results and random sampling (1\% and 3\%) over three dataset sizes, comparing accuracy ($AC$), precision ($PR$), recall ($RE$), F1 score ($F1$), effectiveness ($E$), F1-based effectiveness ($E_{F1}$), and reduction rate ($R$). The best results in each dataset size are bolded.

For small datasets, the proposed CMVML-LSH approach achieves the highest effectiveness (E = 0.796), closely followed by MVML-LSH (E = 0.794) and ML-LSH (E = 0.792). This superior performance can be attributed to the multi-view and multi-level architecture's ability to capture diverse relationship patterns at different granularities. The DM method shows competitive performance (E = 0.791) in small datasets, suggesting that its stratified sampling approach effectively preserves the underlying data distribution even with limited data. Although the original dataset achieves higher raw performance metrics (AC = 0.86), GAIS maintain competitive performance while significantly reducing the dataset size, with reduction rates consistently above 96\%.

In medium-sized datasets, the performance gap between the original and GAIS narrows considerably. The original  achieves an accuracy of 0.945, while MVML-LSH maintains robust performance with an accuracy of 0.895 and effectiveness of 0.868. The DM method demonstrates stable performance (E = 0.872) in medium datasets, outperforming several LSH variants, which suggests that its adaptive similarity threshold mechanism scales well with increasing data size. The multi-view and cross-attention variants show marginal differences in effectiveness (ranging from 0.864 to 0.869), suggesting that the additional complexity of cross-attention mechanisms might not provide substantial benefits. The reduction rates remain consistently high across all proposed methods (96-97\%).

For large datasets, while MVML-LSH achieves the highest effectiveness (E = 0.819) and accuracy (AC = 0.840), its cross-attention variant CMVML-LSH demonstrates superior performance across other key metrics, including precision (PR = 0.715), recall (RE = 0.528) and F1-score (F1 = 0.563). This suggests that the cross-attention mechanism enhances the model's ability to identify more informative instances, leading to better balanced performance. The DM approach shows competitive effectiveness (E = 0.795) while maintaining a high reduction rate (R = 0.98), demonstrating its viability as a computationally efficient alternative. The random sampling baselines consistently underperform compared to the proposed methods across all metrics, validating the importance of structured selection approaches.

Across all dataset sizes, a consistent pattern emerges where the proposed LSH-based methods outperform random sampling approaches in terms of effectiveness while maintaining comparable reduction rates. The DM method demonstrates consistency across different dataset sizes, with effectiveness scores staying within a narrow range, suggesting it provides a reliable baseline regardless of data scale. The multi-view and multi-level architectures show particular strength in preserving the discriminative power of the original dataset, as evidenced by the relatively small drops in precision and recall compared to random sampling.

Table \ref{table:summary_timing_metrics} presents a summary of computational time metrics for GAIS methods across different dataset sizes, comparing graph construction time (GCT), GAT (or GNN) training time (GTT), selection time (IST), and ML training time (MLT).
For small datasets, all proposed methods demonstrate efficient computational performance, with GCT remaining under 2 seconds on average. The DM method shows particularly efficient graph construction (GCT = 0.05s), making it an attractive option for time-sensitive applications with small datasets. While the multi-level variants (ML-LSH, CML-LSH) show slightly higher GCT ($\approx$ 1.9s), they maintain competitive GTT around 4-5 seconds. The IST is negligible across all methods for small datasets.

In medium-sized datasets, the DM method maintains its computational efficiency with the lowest GCT (0.35s), while more complex variants like CMVML-LSH require substantially more time for graph construction (6.0s) and GAT training (33.25s). However, this increased computational cost should be considered alongside the performance benefits discussed earlier. The original ML model training time (2.05s) is notably higher than the ML training time on selected instances (0.10-0.18s), demonstrating the computational advantages of IS.

For large datasets, the DM method demonstrates the highest efficiency with the lowest construction and training times (GCT = 2.55s, GTT = 10.31s), making it particularly suitable for large-scale applications where computational resources are limited. The more complex MVML-LSH and CMVML-LSH methods show significantly higher computational requirements (GCT $>$ 35s, GTT $>$ 80s), though this should be weighed against their enhanced performance metrics. Importantly, the original ML training time (70.64s) is substantially higher than the training time on selected instances (0.28-0.82s), highlighting the significant computational savings achieved through IS, regardless of the chosen method.

\subsection{Comparing Graph Timings}

\begin{figure}[h]
    \centering
    \begin{subfigure}[t]{\linewidth}
        \centering
        \includegraphics[width=\linewidth]{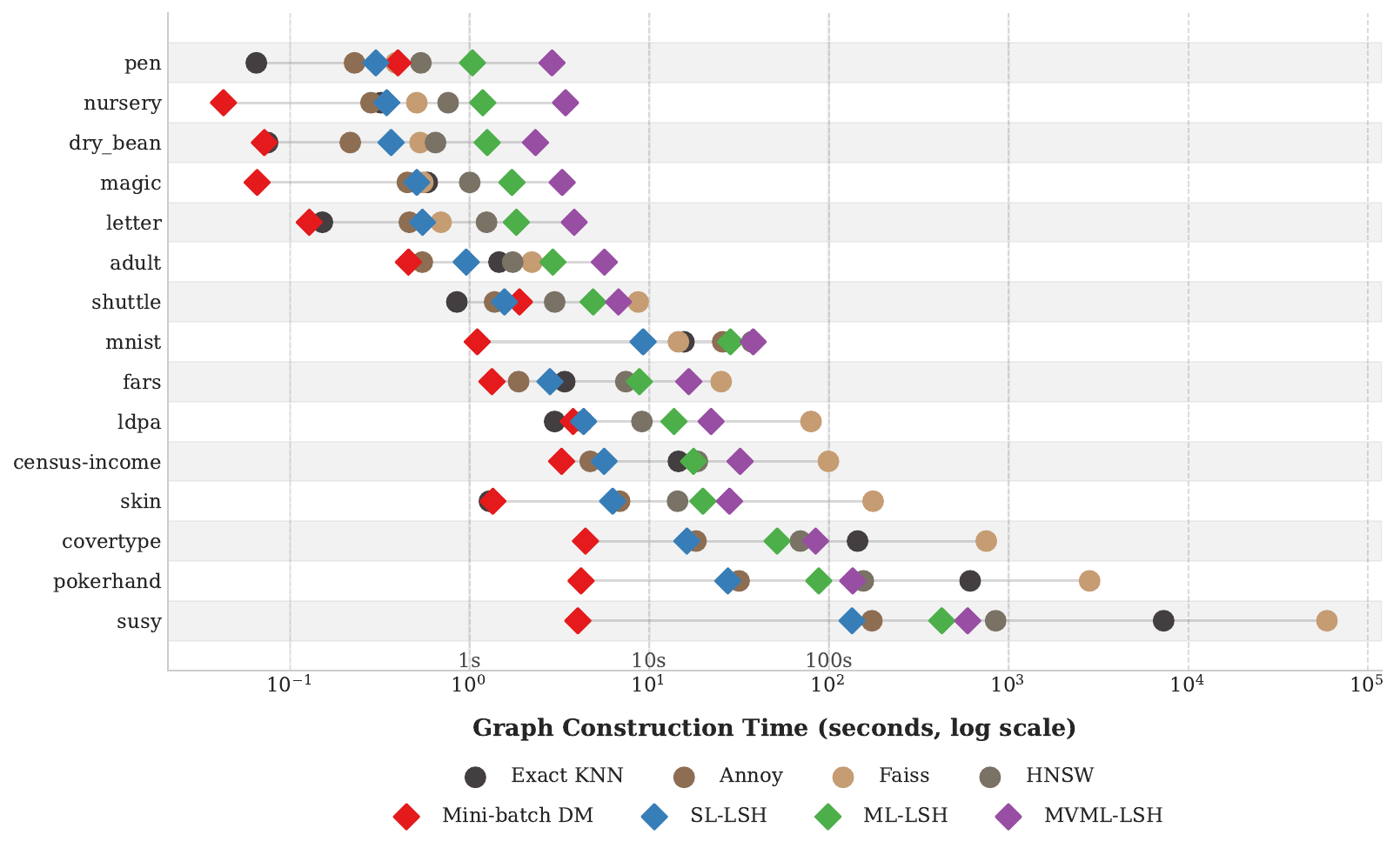}
        \caption{Graph construction time comparison for GAIS variants and KNN-based baselines.}
        \label{fig:graph_construction_time}
    \end{subfigure}
    \begin{subfigure}[t]{\linewidth}
        \centering
        \includegraphics[width=\linewidth]{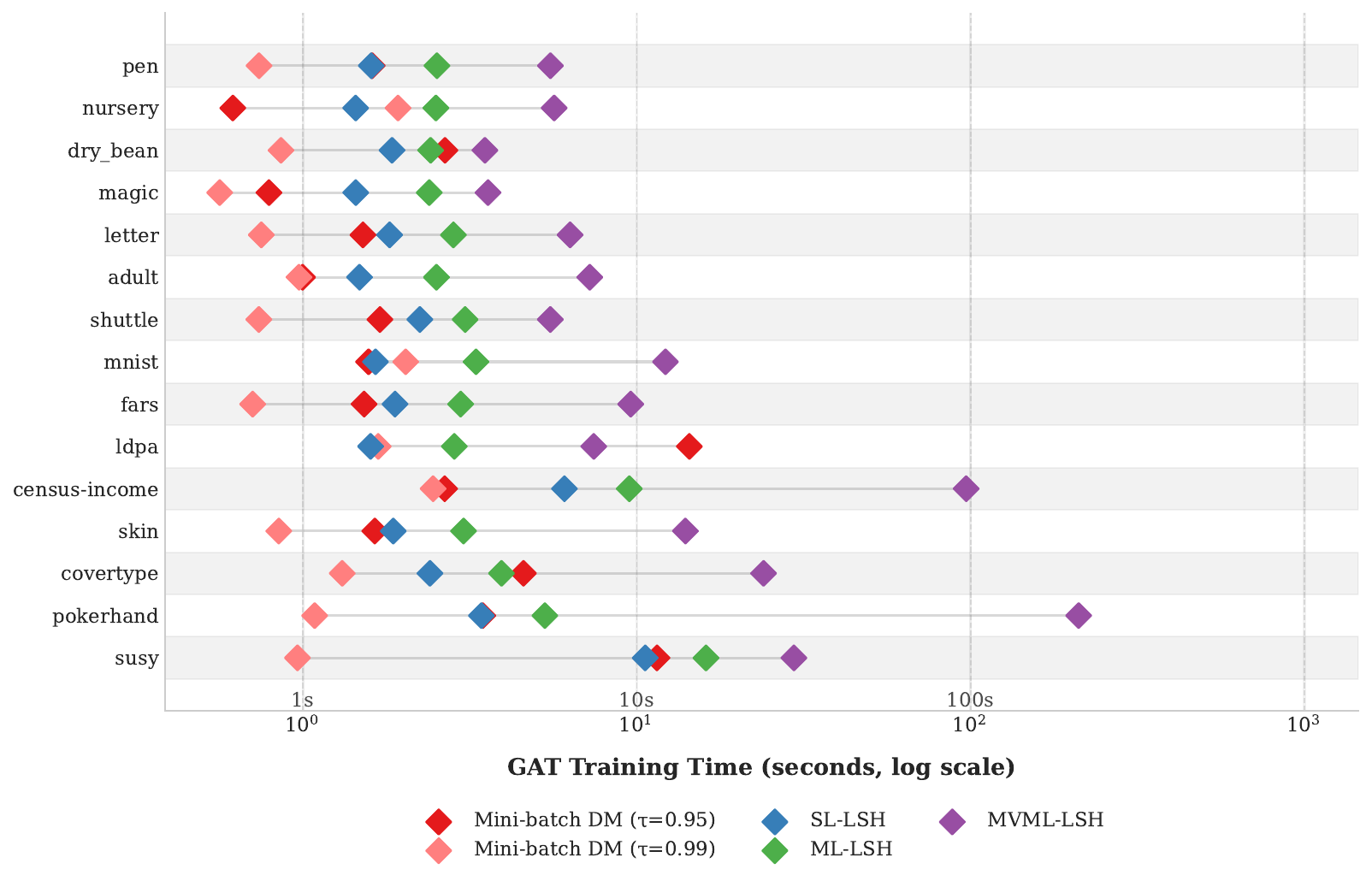}
        \caption{GAT training time comparison for different GAIS variants.}
        \label{fig:gat_training_time}
    \end{subfigure}
    \caption{Scalability analysis of GAIS variants across medium and large-sized datasets: (a) Graph construction time comparison between GAIS variants (using 5 hashtables and projections for LSH methods, 0.95 similarity threshold $\tau$ for DM) and KNN baselines (k=10, n\_trees=10 for Annoy); (b) GAT training time comparison including two DM variants with different similarity thresholds (0.95 and 0.99) to demonstrate the effect of graph sparsity on training efficiency.}
    \label{fig:graph_construction_and_gat_training}
\end{figure}

Figure \ref{fig:graph_construction_time} compares graph construction times between GAIS variants and KNN-based approaches across medium and large datasets. The DM method consistently demonstrates higher efficiency across all dataset sizes, maintaining relatively stable construction times even as dataset size increases. Among the LSH variants, SL-LSH shows the best computational efficiency, followed by ML-LSH, while MVML-LSH exhibits higher construction times due to its more complex multi-view architecture. When compared to KNN-based approaches, all GAIS variants show better scalability, particularly for larger datasets. This is most evident in datasets like SUSY, where exact KNN and FAISS exhibit exponential growth in construction time, while GAIS methods maintain more controlled growth. While approximate KNN methods like Annoy and HNSW show better scaling than exact KNN, they still require significantly more time than the GAIS variants, especially for datasets exceeding 100K instances. These results demonstrate that the proposed GAIS methods, particularly the DM approach, offer a more scalable alternative to traditional KNN-based graph construction methods.

Figure \ref{fig:gat_training_time} illustrates the GAT training times for different GAIS variants, including two versions of the DM method with varying edge sparsity thresholds. The DM method with a 0.99 similarity threshold consistently achieves the fastest training times across all dataset sizes, owing to its highly sparse graph structure retaining only 1\% of the strongest edges. The DM-0.95 version shows higher but still competitive training times, indicating the computational impact of retaining 5\% of the edges. 
Among the LSH variants, SL-LSH is the most efficient, followed by ML-LSH, while MVML-LSH exhibits notably longer training times, particularly for larger datasets like pokerhand and census-income. This escalation in training time correlates with the increasing complexity of the graph structure, from single-level to multi-view architectures.
These results suggest that while more sophisticated graph structures might offer better performance metrics, they come with a significant computational cost in the GAT training phase, highlighting the importance of considering this trade-off when selecting an appropriate method for specific applications.

\subsection{Component Analysis}\label{subsec:component_analysis}

\begin{figure}[htbp]
    \centering
    \includegraphics[width=0.7\linewidth]{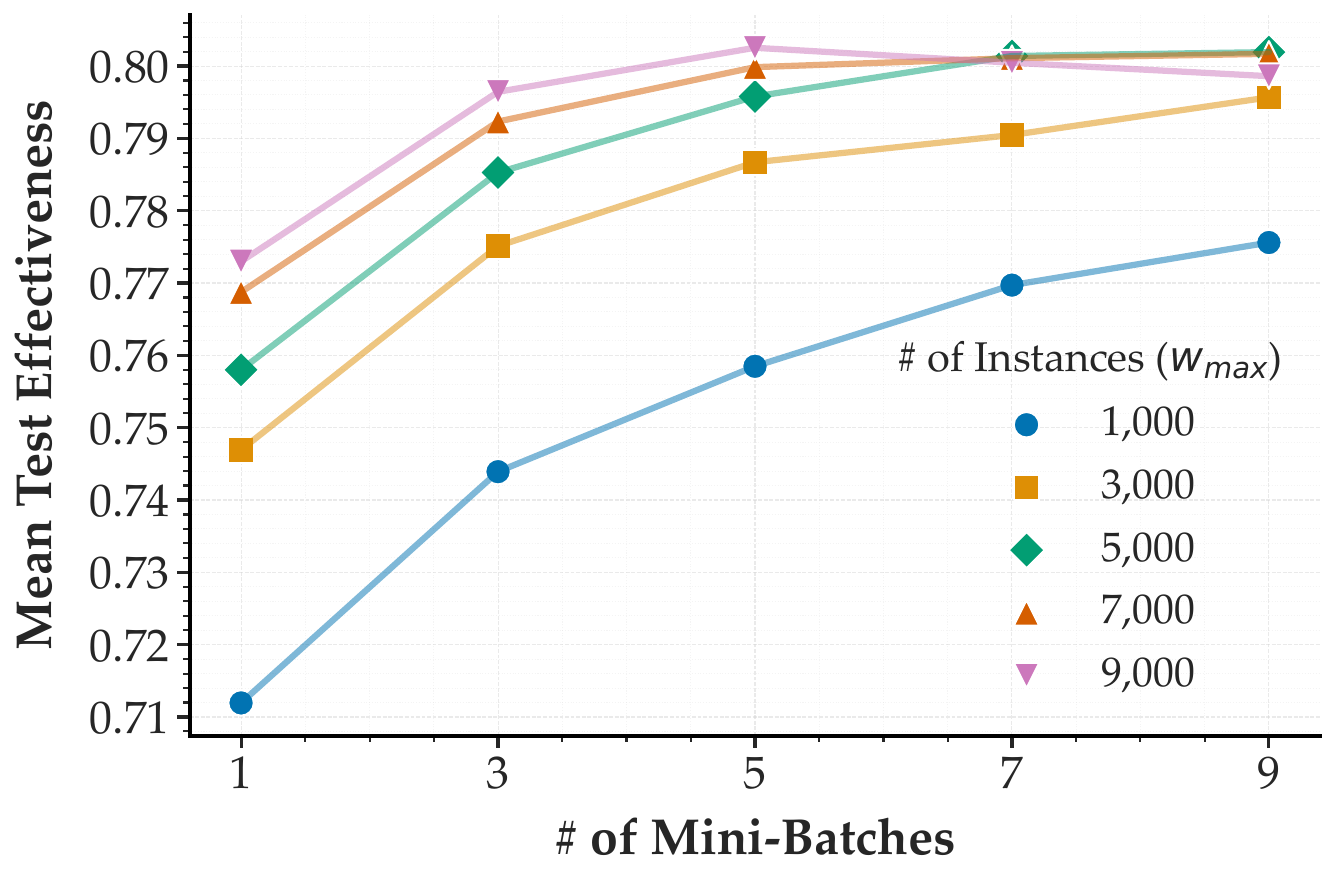}
    \caption{Impact of mini-batch configuration on test effectiveness in the distance-based mini-batch (DM) method for large datasets ($>$100K instances). Results show diminishing returns beyond 7,000 instances per mini-batch and 5-7 mini-batches, providing empirical guidance for setting the computational constraint $w_{\max}$ while maintaining performance.}
    \label{fig:mini_batch_effectiveness}
\end{figure}

Figure \ref{fig:mini_batch_effectiveness} illustrates the relationship between mini-batch configuration and test effectiveness in the DM method for large datasets. We test configurations up to $w_{\max}$ instances per mini-batch to identify natural performance plateaus.
The results demonstrate that both the number of mini-batches and instances per batch have a positive impact on effectiveness, though this improvement shows diminishing returns. As the number of mini-batches increases from 1 to 9, the effectiveness consistently improves across all instance sizes, with the most substantial gains observed in the transition from 1 to 5 mini-batches. Similarly, larger batch sizes generally yield better effectiveness, though the improvement plateaus around $7000$ instances, particularly with higher numbers of mini-batches. This plateau effect is most noticeable when using 7-9 mini-batches, suggesting that beyond these configurations, additional computational resources might not yield proportional performance benefits. These findings provide practical guidance for implementing the DM method, indicating that moderate configurations (5-7 mini-batches with $5000-7000$ instances per batch) might offer an optimal balance between computational efficiency and effectiveness.
The empirical analysis demonstrates that effectiveness plateaus around $7000$ instances per mini-batch, providing direct guidance for $w_{\max}$ selection. This plateau effect validates that computational bounds can be set without sacrificing performance.

\begin{figure}[htbp]
    \centering
    \includegraphics[width=\linewidth]{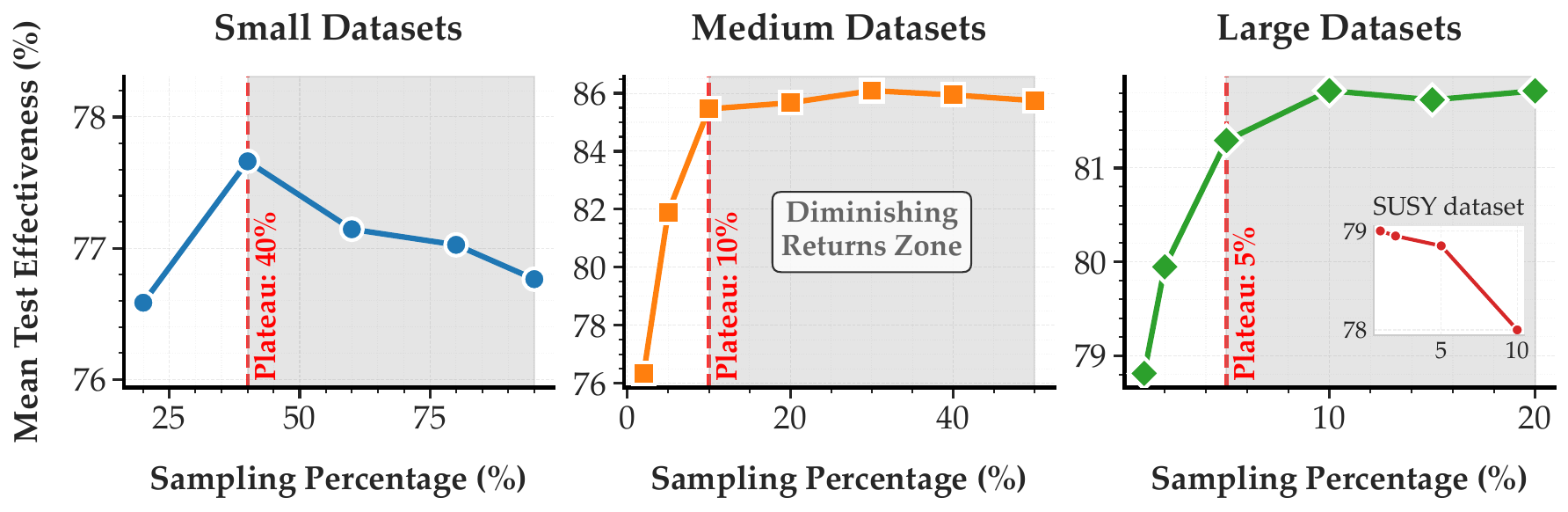}
    \caption{Effectiveness versus sampling percentage by dataset size categories: small ($<$10K instances), medium (10K--100K instances), and large ($>$100K instances). Performance plateaus occur at 5\%, 10\%, and 40\% sampling for large, medium, and small datasets respectively, with the "Diminishing Returns Zone" indicating where additional sampling provides minimal benefit.}
    \label{fig:plateau_analysis}
\end{figure}

Building on these mini-batch insights, we empirically validate our $w_{\text{max}}$ constraint design and address computational scalability concerns by testing performance plateaus across different sampling percentages. This analysis aims to determine the natural point where additional data sampling yields diminishing returns to justify our fixed computational bounds.
Figure \ref{fig:plateau_analysis} demonstrates clear plateau patterns across dataset sizes, with performance stabilizing at low sampling percentages for larger datasets.
Large datasets exhibit performance plateaus at just 5\% sampling, achieving peak effectiveness with a median of 3,601 instances per mini-batch. Importantly, the correlation between dataset size and optimal mini-batch requirements for large datasets is negligible ($r=0.084$, $p=0.875$).
This demonstrates that our approach scales independently of dataset size.
Medium datasets plateau at 10\% sampling with a median of 1,553 instances per mini-batch, while small datasets require 40\% sampling with a median of 160 instances per mini-batch.
These findings support that $w_{\text{max}}$ constraint captures the natural performance ceiling for large datasets while ensuring computational feasibility. The independence of optimal mini-batch size from dataset scale ($r=0.084$ for large datasets) empirically supports our $O(K \times w_{\max}^2)$ complexity analysis.

Table~\ref{tab:lsh_quality} evaluates hash bucket quality across different dataset sizes, comparing the SL- and ML-LSH approaches. The results demonstrate that ML-LSH consistently outperforms SL-LSH across most quality metrics, with particularly notable improvements in purity ($0.81$ vs $0.78$ for small datasets, $0.84$ vs $0.79$ for medium datasets) and $R@5$ performance. Most significantly, ML-LSH exhibits superior scalability characteristics: while both methods experience degraded $R@5$ as dataset size increases, ML-LSH maintains substantially higher recall rates ($0.34$ vs $0.19$ for large datasets), indicating better preservation of local neighborhood structures in large-scale settings. The multi-level approach also produces more computationally efficient bucket distributions, with consistently smaller average bucket sizes ($\bar{B}$) across all dataset categories. The strong correlation values ($\rho$) achieved by ML-LSH, particularly for large datasets ($0.64$ vs $0.59$), confirm that the hierarchical structure effectively captures meaningful similarity relationships while maintaining computational efficiency.

\begin{table}[htbp]
\caption{Comparison of SL- and ML-LSH across dataset sizes. Results averaged across all datasets within each size category. Higher values indicate better performance for all metrics except $\bar{B}$, where smaller values indicate better computational efficiency.}
\label{tab:lsh_quality}
\scriptsize
\begin{threeparttable}
\begin{tabular*}{\columnwidth}{@{\extracolsep\fill}
    >{\raggedright\arraybackslash}p{0.08\columnwidth}|
    >{\centering\arraybackslash}p{0.13\columnwidth}
    >{\centering\arraybackslash}p{0.13\columnwidth}
    >{\centering\arraybackslash}p{0.13\columnwidth}
    >{\centering\arraybackslash}p{0.13\columnwidth}
    >{\centering\arraybackslash}p{0.15\columnwidth}}
\toprule
Method & Sep~$\uparrow$ & Pur~$\uparrow$ & $R@5$~$\uparrow$ & $\rho$~$\uparrow$ & $\bar{B}$~$\downarrow$ \\
\midrule
\multicolumn{6}{c}{\textbf{Small Datasets ($< 10,000$)}} \\
\midrule
SL & 1.23 ± 0.23 & 0.78 ± 0.13 & 0.62 ± 0.29 & 0.45 ± 0.20 & 26.7 ± 24.0 \\
ML & \textbf{1.25 ± 0.25} & \textbf{0.81 ± 0.13} & \textbf{0.66 ± 0.28} & \textbf{0.52 ± 0.20} & \textbf{21.3 ± 22.6} \\
\midrule
\multicolumn{6}{c}{\textbf{Medium Datasets ($10,000 - 100,000$)}} \\
\midrule
SL & 1.21 ± 0.16 & 0.79 ± 0.12 & 0.42 ± 0.33 & 0.53 ± 0.22 & 37.2 ± 18.5 \\
ML & 1.21 ± 0.17 & \textbf{0.84 ± 0.10} & \textbf{0.54 ± 0.36} & \textbf{0.57 ± 0.20} & \textbf{28.8 ± 17.6} \\
\midrule
\multicolumn{6}{c}{\textbf{Large Datasets ($> 100,000$)}} \\
\midrule
SL & 1.32 ± 0.14 & 0.75 ± 0.17 & 0.19 ± 0.20 & 0.59 ± 0.21 & 53.7 ± 28.9 \\
ML & 1.32 ± 0.15 & \textbf{0.79 ± 0.15} & \textbf{0.34 ± 0.23} & \textbf{0.64 ± 0.20} & \textbf{44.5 ± 30.0} \\
\bottomrule
\end{tabular*}
\begin{tablenotes}
\scriptsize
\item Sep = Separation ratio (intra-/inter-bucket cosine similarity); Pur = Purity (class label homogeneity); $R@5$ = Recall@5 (true 5-nearest neighbors in same bucket); $\rho$ = Pearson correlation with cosine similarity; $\bar{B}$ = Average bucket size.
\end{tablenotes}
\end{threeparttable}
\end{table}

\begin{figure}[htbp]
    \centering
    \begin{subfigure}[b]{\columnwidth}
        \centering
        \includegraphics[width=\textwidth]{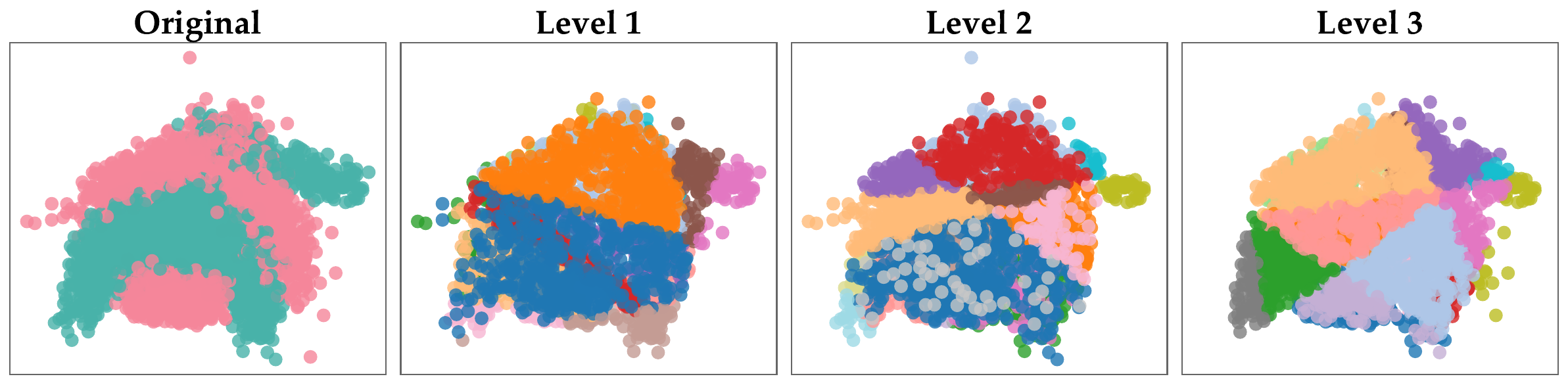}
        \caption{Banana dataset}
        \label{fig:lshbanana}
    \end{subfigure}

    \vspace{0.05cm} 

    \begin{subfigure}[b]{\columnwidth}
        \centering
        \includegraphics[width=\textwidth]{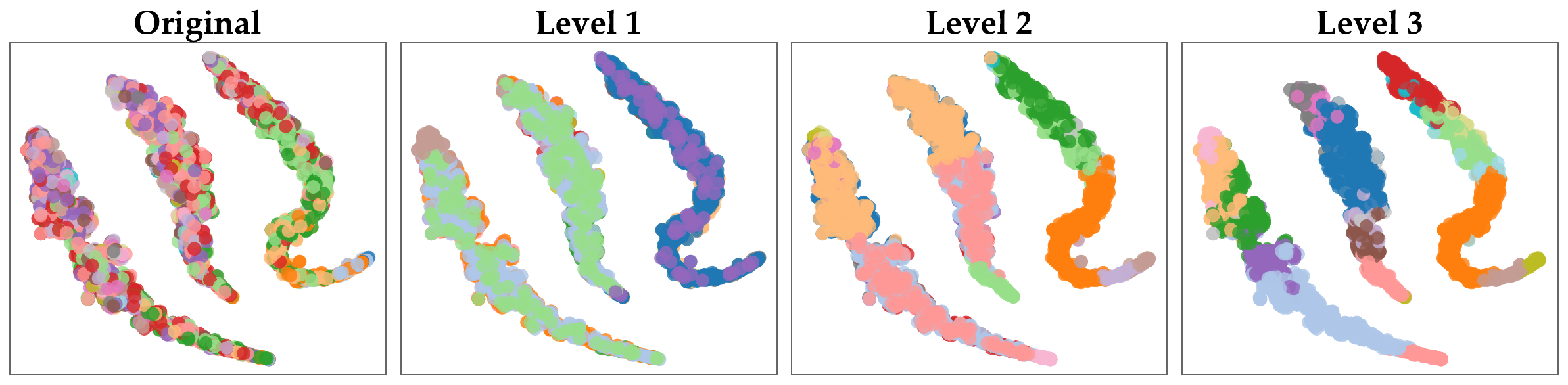}
        \caption{Abalone dataset}
        \label{fig:lsh_abalone}
    \end{subfigure}
    
    \caption{Interpretability of hierarchical Locality-Sensitive Hashing (LSH) bucket assignments via dimensionality reduction. Our Multi-Level LSH (ML-LSH) approach partitions instances into hash buckets at multiple resolution levels to approximate similarity relationships. Each color represents a different hash bucket assignment at each level: Level 1 identifies major structural boundaries (coarse-grained cluster separation), Level 2 refines into coherent sub-regions, and Level 3 captures fine-grained local neighborhoods.}
    \label{fig:lsh_interpretability}
\end{figure}

Figure~\ref{fig:lsh_interpretability} demonstrates the interpretability of our hierarchical LSH approach across two representative datasets with different manifold structures: the \texttt{banana} dataset with crescent-shaped clusters, and the \texttt{abalone} dataset with complex curved manifolds. In both cases, Level 1 identifies major structural boundaries, Level 2 refines these into coherent sub-regions, and Level 3 captures fine-grained local neighborhoods. The consistent ability to discover meaningful structures across different data geometries supports that our multi-level approach effectively approximates semantic similarity relationships while providing interpretable bucket assignments at multiple resolutions.

\begin{figure}[htbp]
    \centering
    \begin{subfigure}[b]{\columnwidth}
        \centering
        \includegraphics[width=\textwidth]{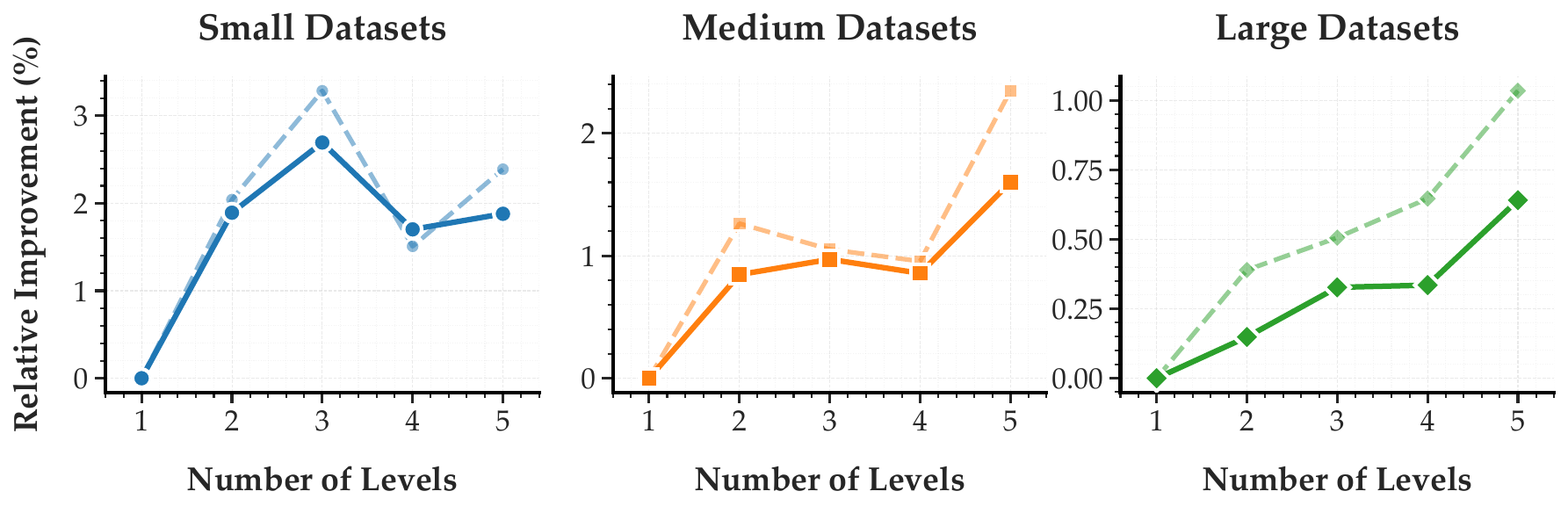}
        \caption{Levels: Hierarchical LSH effectiveness gains}
        \label{fig:levels_lsh}
    \end{subfigure}

    \vspace{0.05cm} 

    \begin{subfigure}[b]{\columnwidth}
        \centering
        \includegraphics[width=\textwidth]{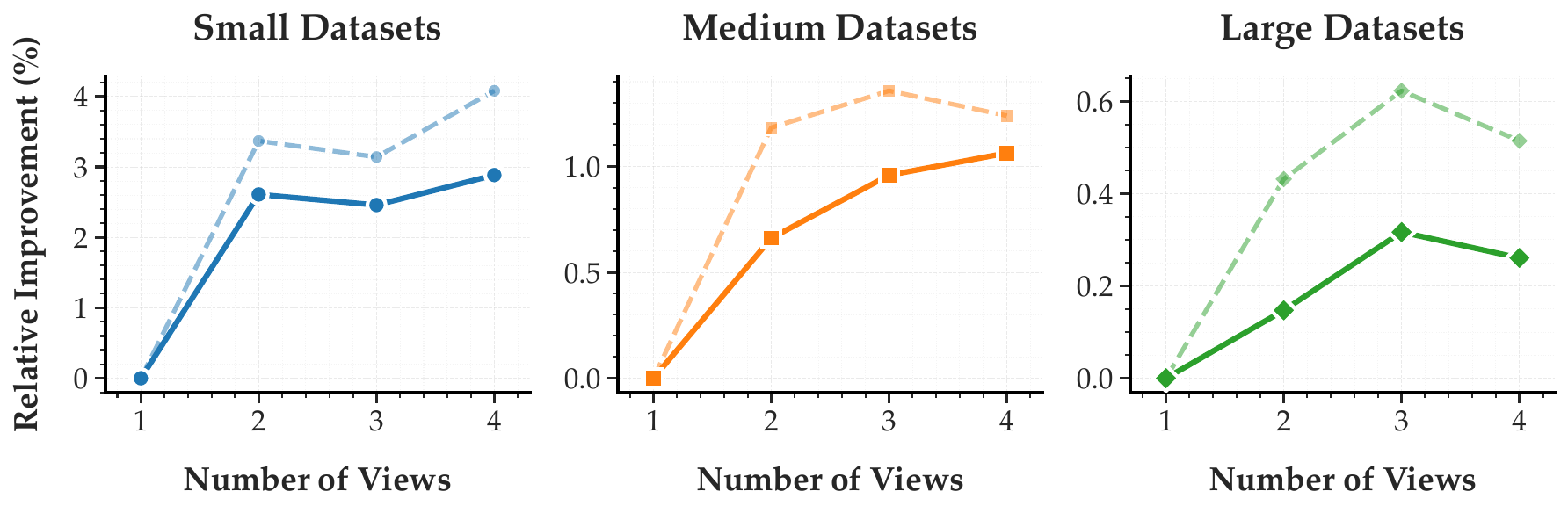}
        \caption{Views: Multi-view LSH effectiveness gains}
        \label{fig:views_lsh}
    \end{subfigure}
    
    \caption{Multi-level and multi-view analysis showing relative improvement (in \%) in test effectiveness for LSH-based graph construction variants. (a) Multi-level analysis: Improvement from using hierarchical levels compared to single-level baseline, across small ($<$10K instances), medium (10K--100K instances), and large ($>$100K instances) datasets. (b) Multi-view analysis: Improvement from using multiple views compared to single-view baseline. Solid lines show average improvements across all datasets in each size category, while dashed lines represent improvements for datasets performing below median effectiveness within each category to highlight challenging datasets.}
    \label{fig:views_levels_lsh}
\end{figure}

Figure \ref{fig:views_levels_lsh} reveals distinct scaling patterns for hierarchical complexity in LSH-based graph construction. For multi-level analysis (Figure \ref{fig:levels_lsh}), small datasets benefit most from increased hierarchy, with Level 3 achieving peak effectiveness improvement (2.69\%), suggesting that fine-grained similarity capture is important when data is limited. This benefit is even more pronounced for lower-performing datasets (3.29\% improvement), indicating that hierarchical structures particularly excel where simpler approaches struggle. Medium datasets show moderate but consistent gains across levels (0.85-1.6\%), with Level 5 providing the highest improvement. Large datasets show smaller relative improvements (0.15-0.64\%) due to already strong baseline performance from sufficient data volume, though the computational overhead of deeper hierarchies may limit additional benefits. For multi-view analysis (Figure \ref{fig:views_lsh}), the pattern is more consistent across dataset sizes: small datasets achieve the largest gains from multiple views (2.88\% peak improvement with 4 views, rising to 4.08\% for lower-performing datasets), as diverse similarity perspectives help overcome limited data representation. Medium and large datasets show progressively smaller but steady improvements, with lower-performing datasets consistently demonstrating larger gains than the overall average. The consistent performance plateau around 3-4 views across all dataset sizes indicates an optimal balance between representational diversity and computational efficiency, beyond which additional views provide marginal returns.

\begin{figure}[htbp]
    \centering
    \includegraphics[width=\linewidth]{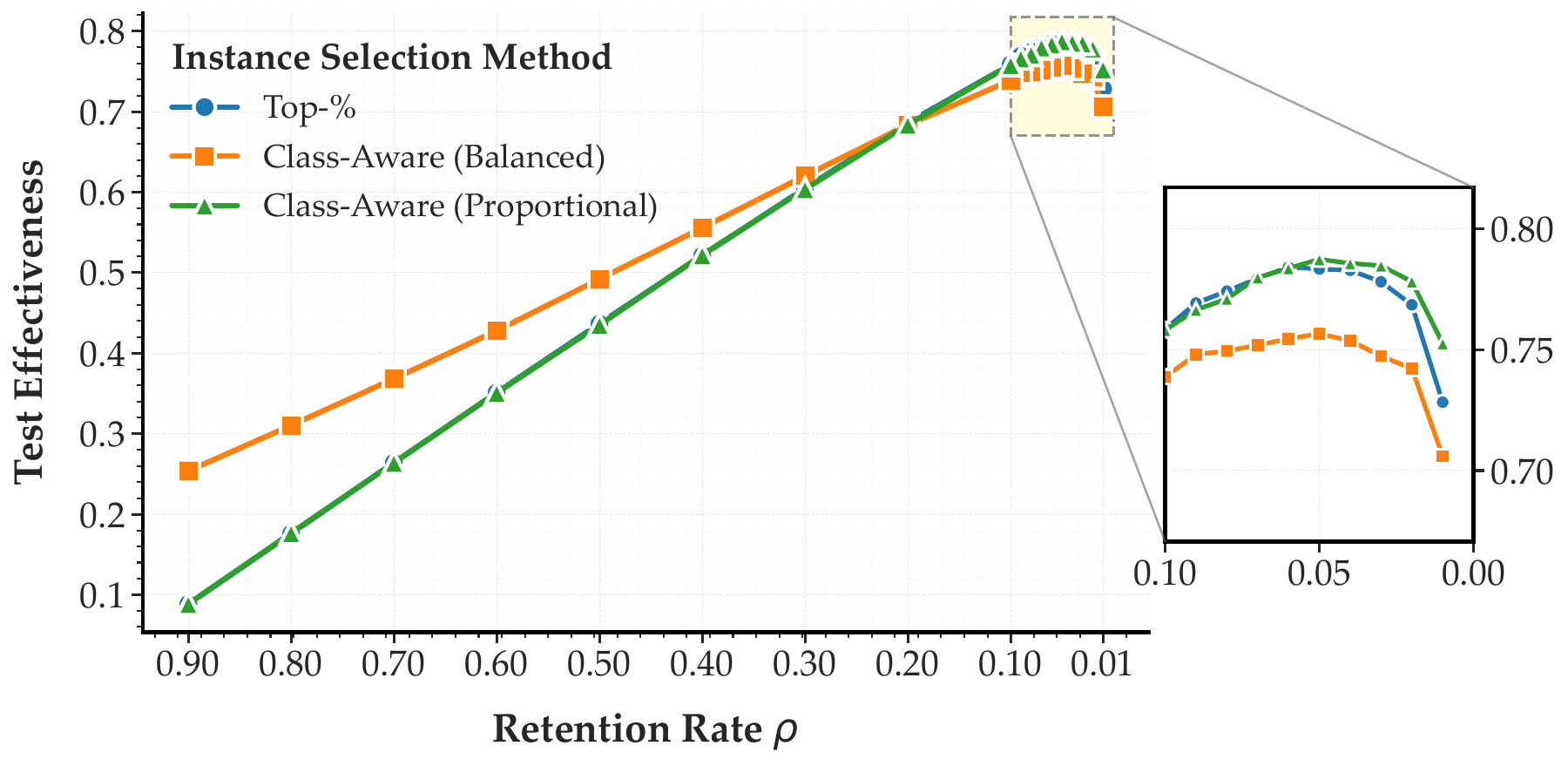}
    \caption{Test effectiveness versus retention rate $\rho$ for three GAIS selection strategies: top-\% (global selection), class-aware balanced (equal per class), and class-aware proportional (maintains class distribution). Inset shows optimal range ($\rho = 0.1$--$0.01$) where proportional strategy performs best, while balanced strategy excels at higher retention rates.}
    \label{fig:confidence_threshold_experiment}
\end{figure}

Figure \ref{fig:confidence_threshold_experiment} reveals how different IS strategies impact test effectiveness across varying retention rates. The balanced class-aware strategy demonstrates better performance at higher retention rates ($\rho = 0.9$–$0.3$), likely due to its ability to maintain equal class representation, which is particularly beneficial when selecting a larger subset of instances. However, this advantage diminishes as the retention rate decreases, with both the global top-percentage and proportional class-aware approaches outperforming the balanced strategy for $\rho < 0.2$. All three strategies exhibit peak performance around the $\rho = 0.06$–$0.04$ range, after which effectiveness begins to decline, most notably in the top-percentage and balanced approaches, while the proportional strategy shows slightly more resilience at lower retention rates. This suggests that while class-aware selection mechanisms are crucial for maintaining model performance when selecting more instances, preserving the original class distribution (proportional strategy) becomes increasingly important as the retention rate decreases toward its optimal level.

\begin{figure}[htbp]
    \centering
    \includegraphics[width=\linewidth]{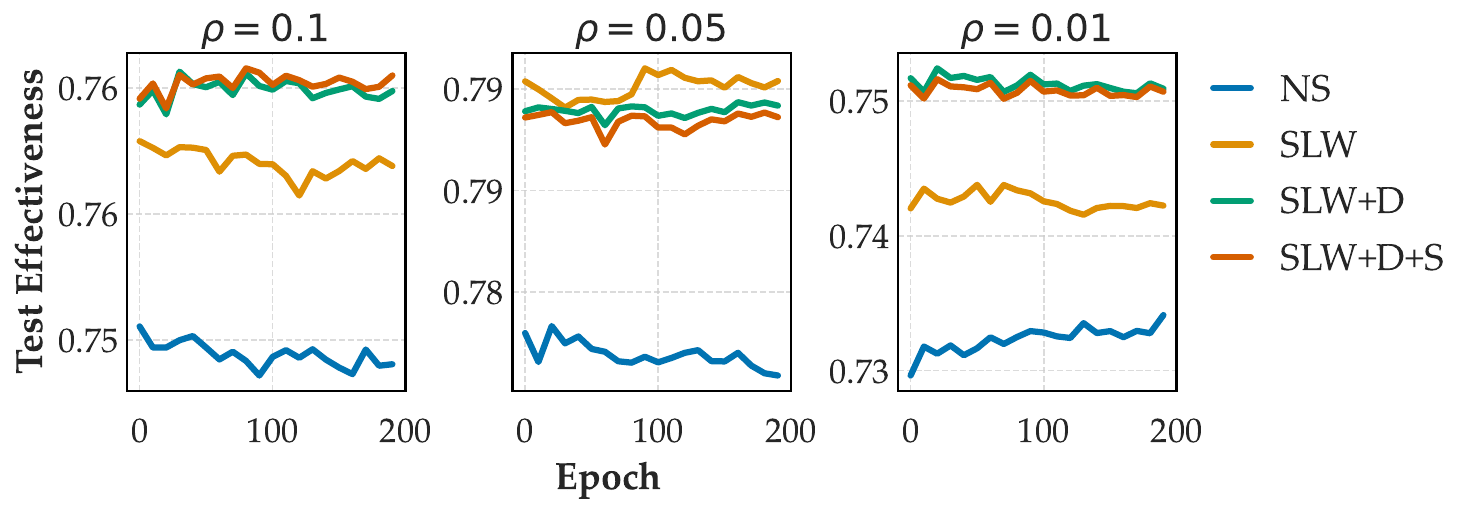}
    \caption{Effect of different normalization strategies on selection effectiveness across three retention ($\rho$) rates: keeping 10\%, 5\%, and 1\% of instances. Four approaches compared: no scaling (NS), layer weighting only (SLW), layer weighting with degree scaling (SLW+D), and layer weighting with degree scaling plus softmax (SLW+D+S). Degree scaling approaches (SLW+D, SLW+D+S) outperform at moderate and highly selective retention ($\rho=0.1 \text{ \& }  0.01$).}
    \label{fig:scaling_effects}
\end{figure}

Figure \ref{fig:scaling_effects} illustrates the impact of different normalization strategies on selection effectiveness across varying retention rates, with results averaged over ten runs using different random seeds to ensure reliability. We examine four approaches: no scaling (NS), layer weighting only (SLW), layer weighting with degree scaling (SLW+D), and layer weighting with both degree scaling and softmax normalization (SLW+D+S). Layer weighting adjusts the contribution of attention weights from different GAT layers, degree scaling normalizes node importance by the square root of their in- and out-degrees, and softmax normalizes the final importance scores to a probability distribution.
At moderate retention rates (0.1, retaining 10\% of instances), approaches that include degree scaling (SLW+D and SLW+D+S) outperform layer weighting alone (SLW), with nearly identical performance (0.76 effectiveness) and notable stability across training epochs. This suggests that degree scaling provides significant benefits when selecting a moderate subset of instances.
At low retention rates (retaining 5\% of instances), layer weighting alone (SLW) achieves slightly higher effectiveness compared to other approaches. This indicates that when selecting a smaller number of instances, the relative importance of different GAT layers becomes more critical in identifying informative instances.
For highly selective rates (retaining only 1\%), the combination of layer weighting and degree scaling (SLW+D) consistently outperforms other approaches (0.75 vs. 0.73 for NS). Adding softmax normalization provides no additional benefit over degree scaling alone.
These findings validate our methodological choice to combine layer weighting and degree scaling in GAIS. While the optimal normalization strategy varies slightly with retention rate, the results clearly show that proper scaling techniques are essential for effective selection, with softmax normalization offering little additional advantage.

\begin{figure}[htbp]
    \centering
    \begin{subfigure}[b]{0.44\columnwidth}
        \centering
        \includegraphics[width=\textwidth]{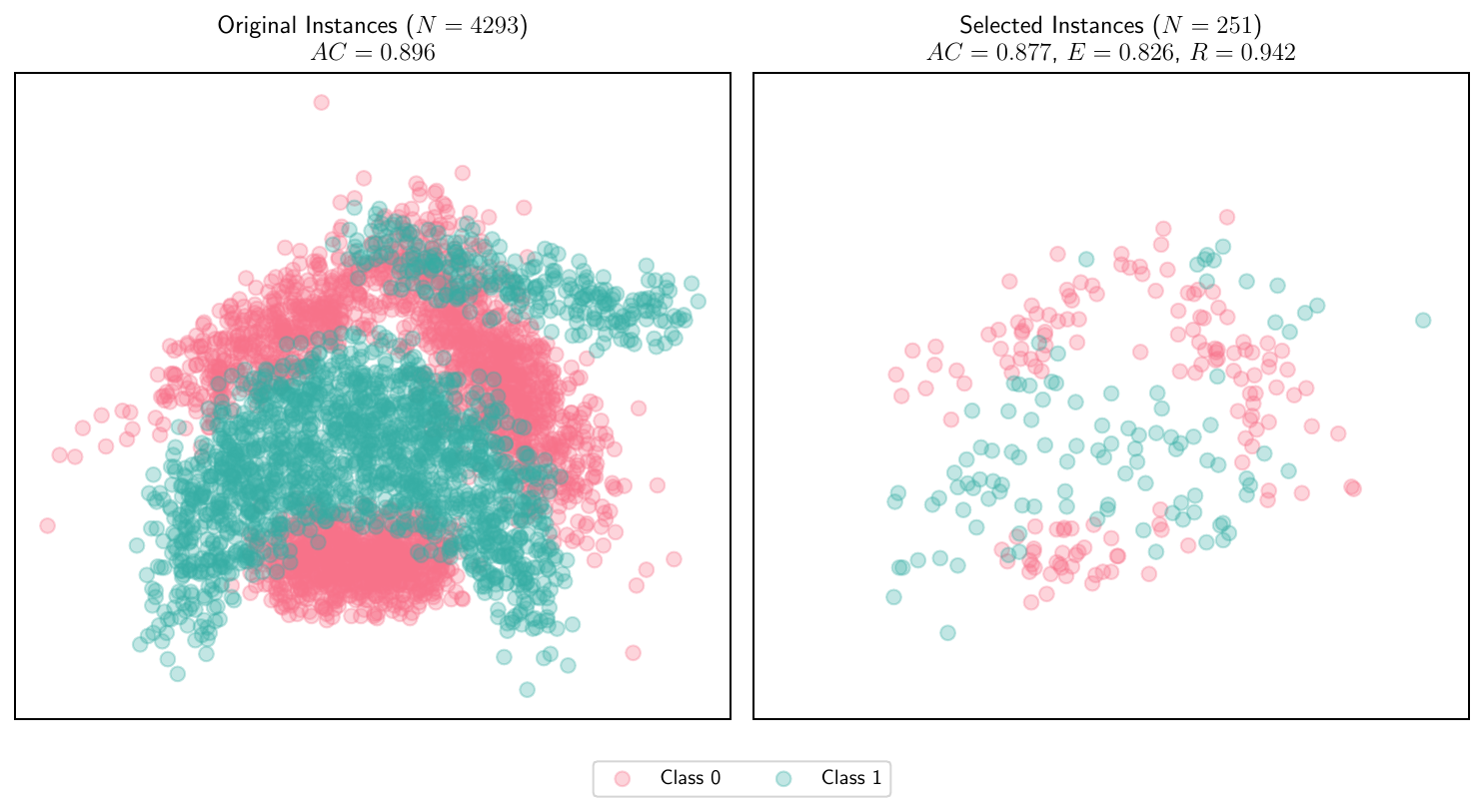}
        \caption{Banana Dataset}
        \label{fig:banana}
    \end{subfigure}
    \begin{subfigure}[b]{0.48\columnwidth}
        \centering
        \includegraphics[width=\textwidth]{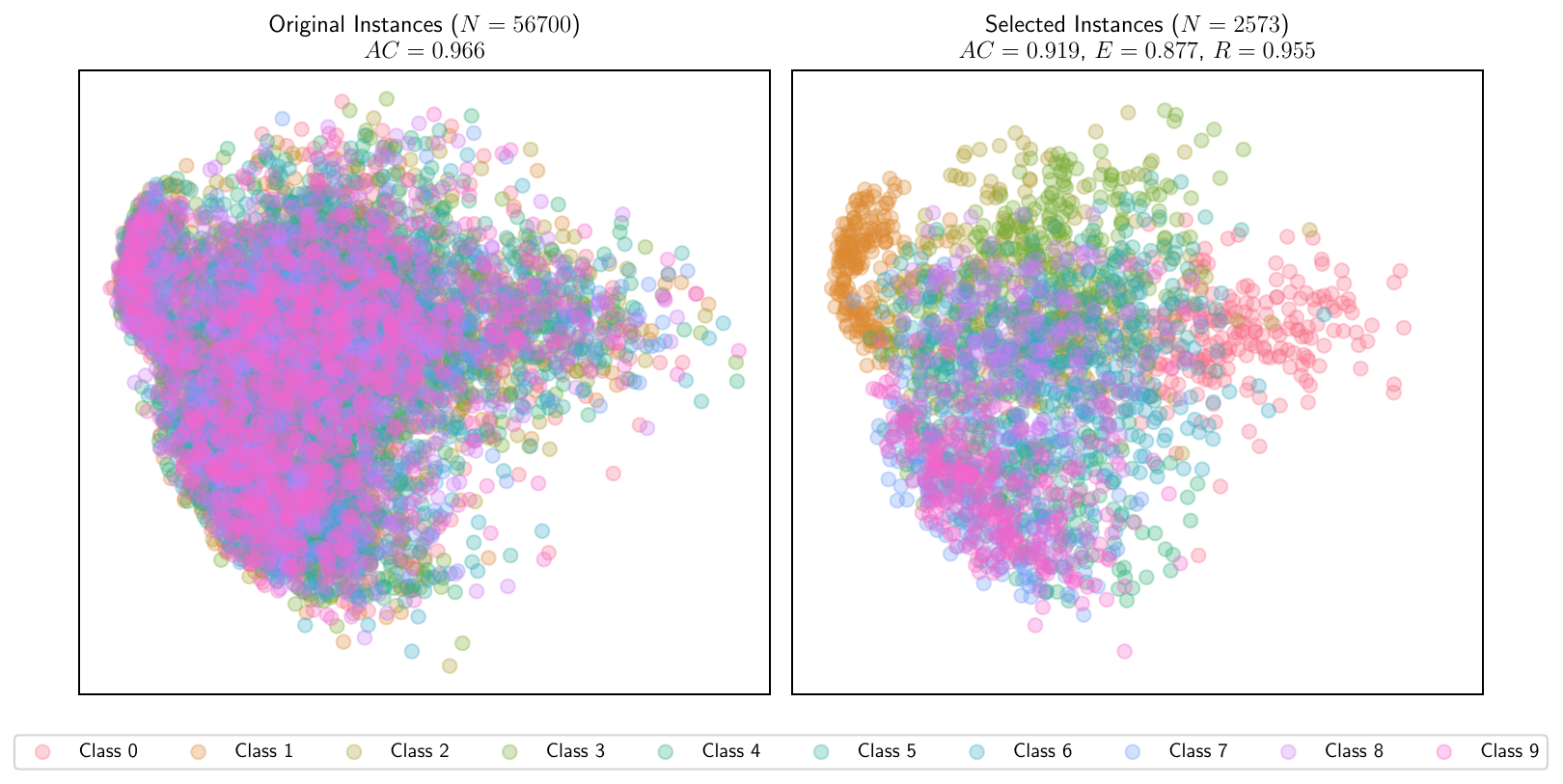}
        \caption{MNIST Dataset}
        \label{fig:mnist}
    \end{subfigure}
    
    \begin{subfigure}[b]{0.48\columnwidth}
        \centering
        \includegraphics[width=\textwidth]{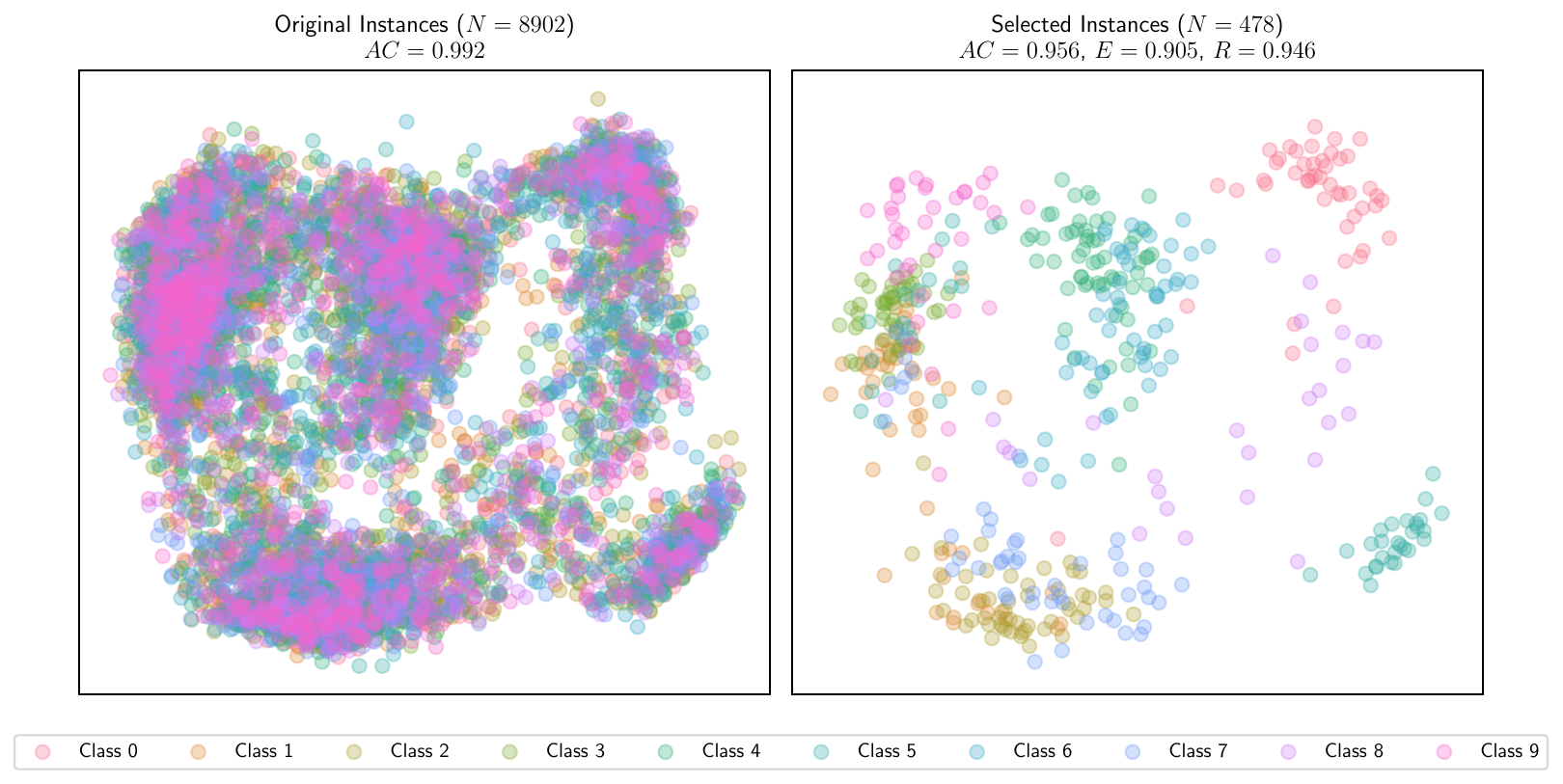}
        \caption{Pen Digits Dataset}
        \label{fig:pen}
    \end{subfigure}
    \begin{subfigure}[b]{0.44\columnwidth}
        \centering
        \includegraphics[width=\textwidth]{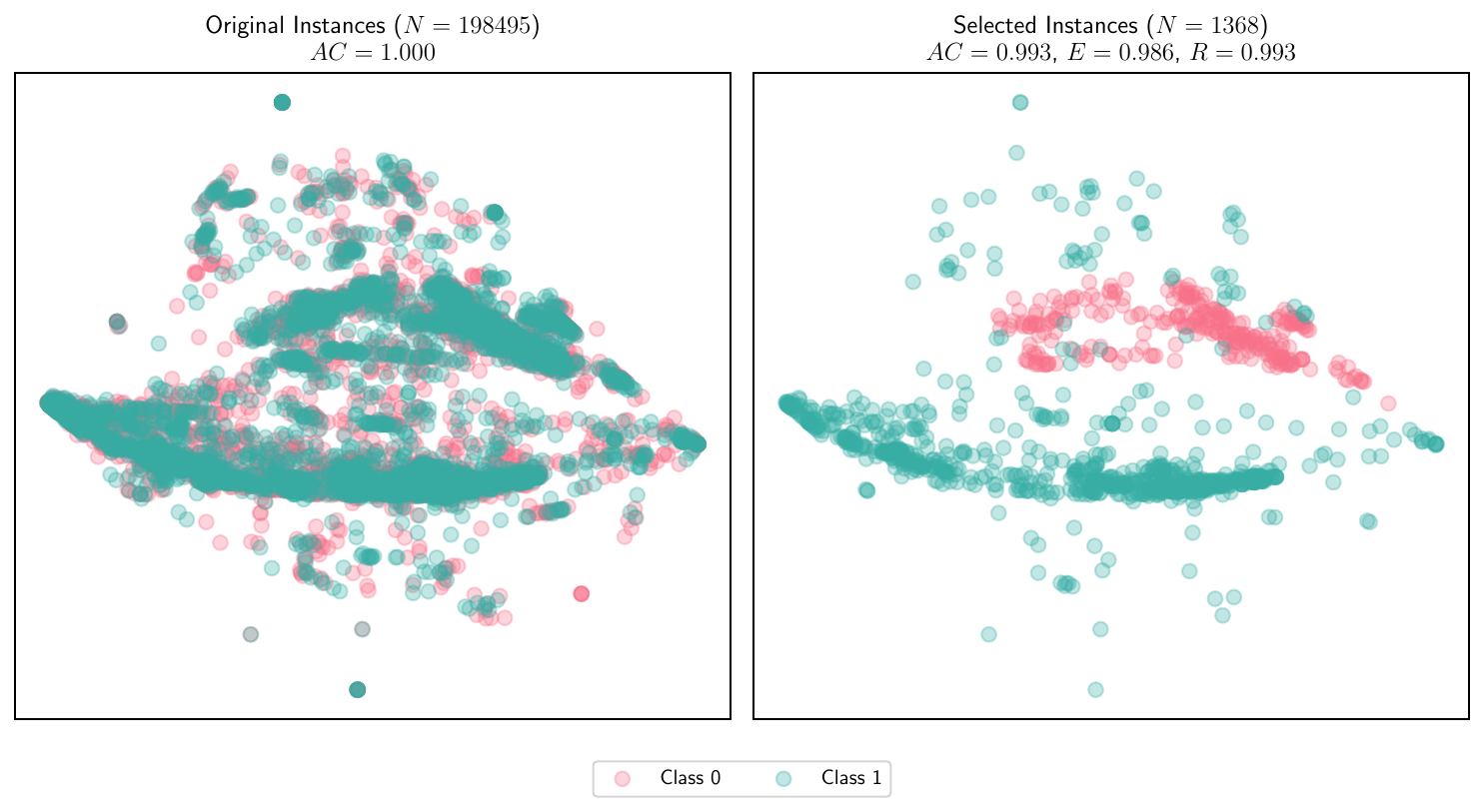}
        \caption{Skin Segmentation Dataset}
        \label{fig:skin}
    \end{subfigure}

    \caption{Qualitative comparison of original (left) versus GAIS-selected (right) instances across four datasets using distance-based mini-batch (DM) method. Different colors represent distinct classes. GAIS maintains class distributions while achieving better separation and reducing redundancy, with selected instances preserving key structural relationships and decision boundaries.}

    \label{fig:qualitative_results}
\end{figure}

Figure \ref{fig:qualitative_results} presents a qualitative comparison between original and selected instances across four datasets. The visualization reveals that while the original datasets exhibit significant class overlap and density in the feature space, the selected instances maintain the inherent class distribution while achieving better class separation. This improved separation suggests that the GAIS effectively identifies instances that are not only representative of their respective classes but also those that define better decision boundaries. Such selective sampling is particularly noticeable in multi-class datasets like MNIST and Pen Digits, where the selected instances appear to preserve the structural relationships between classes while reducing redundancy in densely populated regions.

Table \ref{tab:ablation} demonstrates the incremental value of each proposed GAT enhancement through ablation for two of our approaches: DM and ML-LSH. Head-wise attention amplification (+HeadAmp) provides consistent improvements across both methods, with particularly notable gains in F1 scores (5.2\% for DM, 1.9\% for ML-LSH). This supports our hypothesis that learnable scaling factors enable the model to emphasize attention heads that capture more discriminative relationships. Diversity-promoting attention (+DivAttn) shows method-dependent effects: while it substantially improves DM performance (6.2\% F1 improvement), it slightly reduces ML-LSH F1 scores (-0.4\%). This suggests that the multi-level structure already provides sufficient attention diversity, making explicit diversity-promoting attention less beneficial or potentially redundant. When both are combined (+Both), they consistently deliver the strongest improvements in effectiveness (1.8\% for both methods). This confirms their complementary nature in identifying informative instances. The addition of cross-level attention via Nyström approximation, applicable only to ML-LSH due to its hierarchical structure, provides the most substantial gains (5.6\% F1, 2.1\% effectiveness). This demonstrates that global information exchange across hierarchical levels enhances the model's ability to identify instances with varying structural importance.

\begin{table}[htbp]
\centering
\caption{Ablation study of GAT enhancements for distance-based mini-batching (DM) and multi-level locality-sensitive hashing (ML-LSH) methods. Enhancements include head-wise attention amplification (HeadAmp), diversity-promoting attention (DivAttn), and cross-level attention via Nyström approximation (CrossLevel). Percentages in parentheses show improvement (\textcolor{darkgreen}{$+$}) or decline (\textcolor{red}{$-$}) relative to base GAT.}
\label{tab:ablation}
\scriptsize
\begin{tabular*}{\columnwidth}{@{\extracolsep\fill}
    >{\raggedright\arraybackslash}p{0.07\columnwidth}
    >{\raggedright\arraybackslash}p{0.15\columnwidth}
    >{\centering\arraybackslash}p{0.12\columnwidth}
    >{\centering\arraybackslash}p{0.12\columnwidth}
    >{\centering\arraybackslash}p{0.12\columnwidth}
    >{\centering\arraybackslash}p{0.12\columnwidth}
}
\toprule
Method &  & $AC$ & $F1$ & $E$ & $R$ \\
\midrule
\multirow{4}{*}{DM} & \multicolumn{1}{l}{Base GAT} & \multicolumn{1}{l}{0.827} & \multicolumn{1}{l}{0.655} & \multicolumn{1}{l}{0.795} & \multicolumn{1}{l}{0.959} \\
     & + HeadAmp & 0.835\impr{1.2} & 0.673\impr{5.2} & 0.802\impr{1.1} & 0.958\nimpr{0.1} \\
     & + DivAttn & 0.836\impr{1.3} & \textbf{0.682\impr{6.2}} & 0.804\impr{1.3} & 0.960\impr{0.1} \\
     & + Both & \textbf{0.837\impr{1.5}} & 0.680\impr{6.3} & \textbf{0.807\impr{1.8}} & \textbf{0.963\impr{0.4}} \\
\midrule
\multirow{5}{*}{ML-LSH} & \multicolumn{1}{l}{Base GAT} & \multicolumn{1}{l}{0.824} & \multicolumn{1}{l}{0.652} & \multicolumn{1}{l}{0.794} & \multicolumn{1}{l}{0.961} \\
     & + HeadAmp & 0.832\impr{1.3} & 0.661\impr{1.9} & 0.804\impr{1.5} & \textbf{0.964\impr{0.3}} \\
     & + DivAttn & 0.834\impr{1.3} & 0.657\nimpr{0.4} & 0.805\impr{1.5} & 0.963\impr{0.2} \\
     & + Both & 0.837\impr{1.6} & 0.668\impr{1.3} & 0.806\impr{1.8} & 0.963\impr{0.2} \\
     & + CrossLevel & \textbf{0.838\impr{1.9}} & \textbf{0.676\impr{5.6}} & \textbf{0.809\impr{2.1}} & \textbf{0.964\impr{0.2}} \\
\bottomrule
\end{tabular*}
\end{table}

We compare different GNN architectures within GAIS (specifically the DM method) to thoroughly assess our method's design, where only the GNN component is replaced while maintaining the same graph construction and IS mechanisms, as shown in Table \ref{tab:gnns}. Traditional GNN architectures (GCN, GraphSAGE, GIN \citep{GCN,GraphSAGE,GIN}) achieve comparable effectiveness scores ($0.793-0.795$), while attention-based approaches like GATv2 \citep{GATv2} show similar performance ($0.795$). The Graph Transformer \citep{GraphTransformer} achieves the highest effectiveness among the baseline GNN architectures ($0.796$). However, GAIS with our enhanced GAT architecture consistently outperforms all alternatives across all considered metrics (\(AC=0.837; F1=0.680; E=0.807; R=0.963\)). This supports the importance of our added enhancements which help to better identify informative instances compared to base GNN architectures. Although GNNs can provide viable node importance scores through learned representations, considering attention mechanisms explicitly provides more useful signals for selection.

\begin{table}[htbp]
\centering
\caption{Comparison of different GNN architectures within the GAIS framework using distance-based mini-batching (DM). All methods use identical graph construction and instance selection mechanisms, with only the GNN component varied. Bold indicates best performance.}
\label{tab:gnns}
\scriptsize
\begin{tabular*}{\columnwidth}{@{\extracolsep\fill}
        >{\raggedright\arraybackslash}p{0.25\columnwidth}
        >{\centering\arraybackslash}p{0.12\columnwidth}
        >{\centering\arraybackslash}p{0.12\columnwidth}
        >{\centering\arraybackslash}p{0.12\columnwidth}
        >{\centering\arraybackslash}p{0.12\columnwidth}
    }
    \toprule
    Method & $AC$ & $F1$ & $E$ & $R$ \\
    \midrule
    GCN & 0.829 & 0.675 & 0.795 & 0.958 \\
    GraphSAGE & 0.830 & 0.666 & 0.794 & 0.956 \\
    GIN & 0.826 & 0.672 & 0.793 & 0.958 \\
    GATv2 & 0.827 & 0.666 & 0.795 & 0.962 \\
    Graph Transformer & 0.827 & 0.657 & 0.796 & \textbf{0.963} \\
\midrule
    GAIS (DM) & \textbf{0.837} & \textbf{0.680} & \textbf{0.807} & \textbf{0.963} \\
\bottomrule
\end{tabular*}
\end{table}

\begin{table*}[!t]
\caption{Comparative study of GAIS using distance-based mini-batching (DM) against state-of-the-art IS methods using the effectiveness metric. For studies that did not report this metric, we computed it manually from reported accuracy and reduction values. \textbf{Bold}: highest; \textit{italics}: 2nd highest; \underline{underline}: 3rd highest effectiveness per dataset.}
\label{table:comparative_study}
\scriptsize
\begin{tabular*}{\textwidth}{@{\extracolsep\fill}
    >{\raggedright\arraybackslash}p{0.09\textwidth}|
    >{\centering\arraybackslash}p{0.04\textwidth}
    >{\centering\arraybackslash}p{0.04\textwidth}
    >{\centering\arraybackslash}p{0.04\textwidth}
    >{\centering\arraybackslash}p{0.04\textwidth}
    >{\centering\arraybackslash}p{0.04\textwidth}
    >{\centering\arraybackslash}p{0.04\textwidth}
    >{\centering\arraybackslash}p{0.04\textwidth}
    >{\centering\arraybackslash}p{0.04\textwidth}
    >{\centering\arraybackslash}p{0.04\textwidth}
    >{\centering\arraybackslash}p{0.06\textwidth}
    >{\centering\arraybackslash}p{0.06\textwidth}
    >{\centering\arraybackslash}p{0.05\textwidth}
    >{\centering\arraybackslash}p{0.04\textwidth}
}
\toprule
Dataset & RIS 1 & BDIS & CNN & ENN & ICF & LDIS & CDIS & GDIS & EGDIS & EIS-AS & CIS-BE & S2EJR-2 & \textbf{GAIS\textsuperscript{\dag}} \\
\midrule
abalone & --- & --- & 0.016 & 0.192 & \textit{0.209} & 0.151 & 0.150 & \textbf{0.228} & 0.111 & --- & --- & --- & \underline{0.204} \\
adult & \underline{0.355} & --- & --- & --- & --- & --- & --- & --- & --- & --- & \textit{0.821} & --- & \textbf{0.830} \\
banana & --- & \textit{0.823} & 0.631 & 0.103 & \underline{0.819} & 0.759 & 0.709 & 0.761 & 0.804 & --- & --- & --- & \textbf{0.852} \\
car & --- & --- & \textit{0.687} & 0.061 & 0.327 & 0.640 & 0.662 & 0.602 & \underline{0.665} & --- & --- & --- & \textbf{0.761} \\
census-income & --- & --- & --- & --- & --- & 0.796 & 0.801 & \underline{0.809} & \textit{0.836} & --- & --- & --- & \textbf{0.942} \\
chess & --- & --- & \textit{0.751} & 0.033 & 0.596 & 0.728 & \underline{0.728} & 0.697 & 0.726 & --- & --- & --- & \textbf{0.916} \\
coil2000 & 0.292 & --- & --- & --- & --- & \textit{0.816} & \underline{0.806} & 0.752 & 0.792 & --- & --- & --- & \textbf{0.939} \\
contraceptive & 0.123 & --- & 0.119 & 0.254 & 0.350 & \textit{0.437} & \underline{0.430} & 0.422 & 0.414 & --- & --- & --- & \textbf{0.478} \\
covertype & --- & --- & --- & --- & --- & --- & --- & --- & --- & \underline{0.585} & \textit{0.620} & --- & \textbf{0.795} \\
diabetes & --- & --- & --- & --- & --- & --- & --- & --- & --- & --- & --- & \textit{0.412} & \textbf{0.681} \\
fars & --- & --- & --- & --- & --- & 0.597 & 0.597 & \underline{0.644} & \textit{0.670} & --- & --- & --- & \textbf{0.764} \\
german & --- & --- & 0.306 & 0.203 & \underline{0.592} & 0.546 & 0.501 & 0.533 & \textit{0.608} & --- & --- & --- & \textbf{0.709} \\
heart & \textit{0.380} & --- & --- & --- & --- & --- & --- & --- & --- & --- & --- & --- & \textbf{0.782} \\
ldpa & --- & --- & --- & --- & --- & 0.580 & 0.595 & \underline{0.625} & \textit{0.626} & --- & --- & --- & \textbf{0.706} \\
letter & --- & \textit{0.752} & --- & --- & --- & --- & --- & --- & --- & --- & --- & \underline{0.705} & \textbf{0.768} \\
magic & --- & \textit{0.722} & --- & --- & --- & --- & --- & --- & --- & --- & --- & --- & \textbf{0.826} \\
nursery & --- & --- & --- & --- & --- & 0.375 & 0.540 & \textit{0.572} & \underline{0.560} & --- & --- & --- & \textbf{0.872} \\
opt-digits & --- & 0.877 & \underline{0.883} & 0.012 & \textbf{0.899} & 0.755 & 0.753 & 0.821 & 0.826 & --- & --- & 0.862 & \textit{0.889} \\
page\_blocks & --- & --- & 0.794 & 0.039 & \textit{0.880} & 0.809 & 0.803 & 0.821 & 0.834 & --- & --- & \underline{0.855} & \textbf{0.934} \\
pen & --- & \textbf{0.921} & --- & --- & --- & --- & --- & --- & --- & --- & --- & --- & \textit{0.917} \\
phoneme & --- & --- & 0.615 & 0.098 & \underline{0.729} & 0.716 & 0.723 & 0.683 & \textit{0.730} & --- & --- & --- & \textbf{0.782} \\
pokerhand & --- & --- & --- & --- & --- & --- & --- & --- & --- & \textit{0.443} & --- & --- & \textbf{0.577} \\
ringnorm & --- & \underline{0.758} & 0.621 & 0.034 & \textit{0.815} & 0.492 & 0.519 & 0.658 & 0.504 & --- & --- & --- & \textbf{0.882} \\
satellite & 0.132 & 0.791 & 0.691 & 0.082 & \textit{0.796} & \underline{0.796} & 0.784 & 0.734 & 0.762 & --- & --- & --- & \textbf{0.815} \\
segment & \underline{0.820} & --- & \textit{0.823} & 0.043 & 0.754 & 0.792 & 0.808 & 0.744 & 0.759 & --- & --- & 0.743 & \textbf{0.843} \\
shuttle & --- & \textit{0.953} & --- & --- & --- & --- & --- & --- & --- & --- & --- & --- & \textbf{0.987} \\
skin & --- & 0.958 & --- & --- & --- & 0.680 & 0.431 & \underline{0.973} & \textit{0.973} & --- & --- & --- & \textbf{0.990} \\
spambase & --- & --- & 0.641 & 0.095 & \textit{0.704} & 0.678 & 0.688 & 0.667 & \underline{0.700} & --- & --- & 0.490 & \textbf{0.893} \\
splice & --- & --- & 0.456 & 0.160 & 0.480 & \textit{0.578} & \underline{0.557} & 0.527 & 0.519 & --- & --- & --- & \textbf{0.843} \\
susy & --- & --- & --- & --- & --- & --- & --- & --- & --- & \textit{0.776} & --- & --- & \textbf{0.789} \\
texture & --- & \textbf{0.910} & \textit{0.873} & 0.012 & 0.842 & 0.846 & 0.832 & 0.810 & 0.825 & --- & --- & --- & \underline{0.859} \\
thyroid & --- & 0.823 & 0.691 & 0.057 & \textit{0.912} & 0.777 & 0.770 & 0.826 & \underline{0.852} & --- & --- & --- & \textbf{0.960} \\
titanic & \textit{0.122} & --- & --- & --- & --- & --- & --- & --- & --- & --- & --- & --- & \textbf{0.750} \\
twonorm & --- & \underline{0.910} & 0.743 & 0.179 & 0.868 & 0.801 & 0.794 & 0.878 & \textit{0.915} & --- & --- & --- & \textbf{0.976} \\
yeast & 0.151 & --- & 0.153 & 0.263 & \underline{0.467} & 0.401 & 0.397 & \textit{0.484} & 0.423 & --- & --- & --- & \textbf{0.524} \\
\bottomrule
\end{tabular*}
\vspace{0.5em}
\noindent\makebox[\textwidth]{%
    \footnotesize\textbf{Bold}: highest; \textit{italics}: 2nd highest; \underline{underline}: 3rd highest value in each row
    \hfill
    \textsuperscript{\dag} distance-based mini-batching (DM) GAIS
}
RIS 1: \cite{Cavalcanti2020};
BDIS: \cite{Chen2022b};
CNN---EGDIS: \cite{Malhat2020};
EIS-AS: \cite{Gong2021};
CIS-BE: \cite{Moran2022};
S2EJR-2: \cite{SanchezFernandez2024}
\vspace{0.5em}
\end{table*}

\subsection{Comparative Study}

The comparative analysis in Table \ref{table:comparative_study} demonstrates GAIS's effectiveness against state-of-the-art IS methods. The mini-batch DM variant of GAIS achieves superior performance in 31 out of 35 datasets, with particularly significant improvements in both binary and multi-class classification problems. In large, complex datasets such as census-income (0.942 vs EGDIS's 0.836) and coil2000 (0.939 vs LDIS's 0.816), GAIS shows substantial improvements over existing methods. For highly discriminative datasets like skin (0.990 vs GDIS's 0.973) and thyroid (0.960 vs ICF's 0.912), the improvements are more modest but consistent. GAIS demonstrates particularly strong scalability on massive datasets, achieving strong performance on SUSY (0.789), covertype (0.795), and pokerhand (0.577) which contain millions of instances. In datasets with high class imbalance or complex feature interactions, such as yeast (0.524) and contraceptive (0.478), GAIS maintains its performance advantage over both traditional methods like CNN (0.153, 0.119) and more recent approaches like EGDIS (0.423, 0.414).
While there are potential variations in experimental configurations across studies (e.g., random seeds, data splitting strategies), the consistent higher performance across such diverse datasets suggests that GAIS's fundamental advantages --- its ability to capture multi-scale relationships through graph construction and adaptively weight instance importance through attention mechanisms --- enable it to better identify truly informative instances compared to both density-based methods like EGDIS and learning-based clustering approaches like CIS-BE. Unlike traditional methods that rely on fixed neighborhood metrics or static cluster assignments, GAIS's architecture allows it to learn and leverage both local and global data patterns.


\subsection{Limitations}

Despite GAIS's demonstrated effectiveness, several important limitations warrant discussion. The framework currently lacks theoretical guarantees regarding the optimality of selected instances and the preservation of crucial data characteristics, relying primarily on empirical validation. The multi-level and multi-view architectures, while effective, introduce additional hyperparameters that require tuning through computationally expensive Bayesian optimization, potentially limiting deployment in resource-constrained environments. Also, the attention-based importance scoring mechanism, though interpretable, is constrained by the graph structure's locality, potentially missing important global relationships that exist beyond immediate neighborhoods. This locality constraint becomes particularly relevant in high-dimensional spaces where the curse of dimensionality can affect the quality of graph construction. The framework also assumes that the input features are sufficiently informative for meaningful graph construction, as it operates directly in the original feature space without learning more complex representations. These limitations suggest valuable directions for future research, particularly in developing theoretical foundations and more efficient hyperparameter optimization strategies.

\section{Conclusions}\label{sec:conclusions}

This study introduces GAIS, showing that graph attention mechanisms can effectively capture the intrinsic relationships between instances to guide the selection process. Our experiments across 39 diverse datasets reveal several insights about the relationship between data characteristics and selection effectiveness. The consistent outperformance of MVML-LSH on complex, high-dimensional datasets suggests that viewing instance relationships through multiple granularities and distance metrics offers a more detailed understanding of data importance. In contrast, the competitive performance of the simpler DM approach, particularly in large-scale scenarios, shows that strategic sampling with adaptive thresholds can match the results of more complex architectures when computing resources are limited.

The analysis shows an interesting pattern: GAIS performs relatively better than existing methods when dealing with datasets that have complex feature interactions and moderate class imbalance. This suggests that graph-based approaches are effective at retaining the important decision boundaries. Our experiments with cross-attention mechanisms shows that allowing information to flow between different views and levels can capture subtle but important relationships, though this benefit must be weighed against the increased computational costs.

The success of attention-based importance scoring suggests that local structural properties, when properly aggregated through multiple views and levels, can effectively identify instances important for maintaining decision boundaries. This challenges the traditional notion that IS necessarily requires global distance computations, and opens new directions to make preprocessing more efficient.
Since graph-based methods show promise for modeling relationships across different data modalities, GAIS could be useful for selecting data in image and text domains, where data preprocessing remains challenging.

Future research should look into developing theoretical foundations for the relationship between attention patterns and instance importance, particularly for preserving decision boundaries. Also, exploring adaptive strategies for configuring views and levels based on dataset characteristics could make GAIS more practical across different fields.

\section*{Acknowledgements}
The authors would like to thank Miryam de Lhoneux for her valuable feedback and insightful comments on the manuscript.



\end{document}